%% file: ms.tex
\documentclass{article} %

\usepackage[a4paper, margin=1in]{geometry}

\usepackage{tikz,graphicx}

\usepackage{resizegather}

\usepackage{algorithm}
\usepackage[noend]{algcompatible}
\usepackage{subcaption}
\usepackage{url}
\usepackage{booktabs}

\usepackage{xspace}
\usepackage{balance}

\usepackage{amsmath}

\usepackage{titling}
\predate{}
\postdate{}
\date{}
\preauthor{}
\postauthor{}
\author{}

\newcommand{\approach}{\textsc{ST-Discovery}\xspace}
\newcommand{\landau}{\mathcal{O}}

\title{Mining Topological Dependencies of Recurrent Congestion in Road Networks} 
\begin{document}
\maketitle
\begin{center}
Nicolas Tempelmeier$^1$, Udo Feuerhake$^2$, Oskar Wage$^2$, Elena Demidova$^{1,3}$ \\[0.2cm]
$^{1}$L3S Research Center, Leibniz University Hannover, Germany;\\
$^{2}$Institute of Cartography and Geoinformatics, Leibniz University Hannover, Germany;\\
$^{3}$Data Science \& Intelligent Systems Group (DSIS), University of Bonn, Germany\\[0.2cm]
\{tempelmeier, demidova\}@L3S.de \{Udo.Feuerhake, Oskar.Wage\}@ikg.uni-hannover.de\\
\end{center}

\input{0.1_abstract.tex}

\input{1_introduction.tex}

\input{2_relatedwork.tex}

\input{3_problem.tex}

\input{4_approach.tex}

\input{5_datasets.tex}

\input{6.0_evaluation.tex}

\input{7_conclusion.tex}

\section*{Acknowledgments}
This work is partially funded by the BMBF and the BMWi, Germany under the projects  
"Data4UrbanMobility" (grant ID 02K15A040), "USEfUL" (grant ID 03SF0547),  "CampaNeo" (grant ID 01MD19007B), "d-E-mand" (grant ID 01ME19009B), the European Commission (EU H2020, ``smashHit", grant-ID 871477) as well as by the research initiatives "Mobiler Mensch" and "Urbane Logistik Hannover".

\bibliographystyle{amsplain}
\bibliography{ref}

\end{document}

%% file: 0.1_abstract.tex
\abstract{
The discovery of spatio-temporal dependencies within urban road networks that cause Recurrent Congestion (RC) patterns is crucial for numerous real-world applications, including urban planning and scheduling of public transportation services.  
While most existing studies investigate temporal patterns of RC phenomena, the influence of the road network topology on RC is often overlooked.
This article proposes the \approach algorithm, a novel unsupervised spatio-temporal data mining algorithm that facilitates the effective data-driven discovery of RC dependencies induced by the road network topology using real-world traffic data.
We factor out regularly reoccurring traffic phenomena, such as rush hours, mainly induced by the daytime, by modelling and systematically exploiting temporal traffic load outliers.  
We present an algorithm that first constructs connected subgraphs of the road network based on the traffic speed outliers.
Second, the algorithm identifies pairs of subgraphs that indicate spatio-temporal correlations in their traffic load behaviour to identify topological dependencies within the road network. 
Finally, we rank the identified subgraph pairs based on the dependency score determined by our algorithm.
Our experimental results demonstrate that \approach can effectively reveal topological dependencies in urban road networks.
}

%% file: 1_introduction.tex
\section{Introduction}
\label{sec:introduction}

Urban road networks possess complex interdependencies that can become apparent during congestion events \cite{DBLP:journals/epjds/GuoZFTZLLH19}.
Established traffic research distinguishes between Recurrent Congestion (RC), e.g., rush hours, and Non-Recurrent Congestion events, e.g., accidents \cite{doi:10.3141/1867-08}.
This article aims to identify topological dependencies within the road network that may cause RC phenomena, henceforth called \emph{structural dependencies}.
Such dependencies are often not well understood and can become apparent only under real traffic load and can cause co-occurring RC patterns in the road network. 
Therefore, understanding topological dependencies in urban road networks is crucial for many real-world applications, including city planning, traffic management, and public transportation services.

\begin{figure}
    \centering
       \begin{tikzpicture}
        \node[inner sep=0pt] (picture) {\includegraphics[trim=0 20 0 20, clip, width=0.6\textwidth]{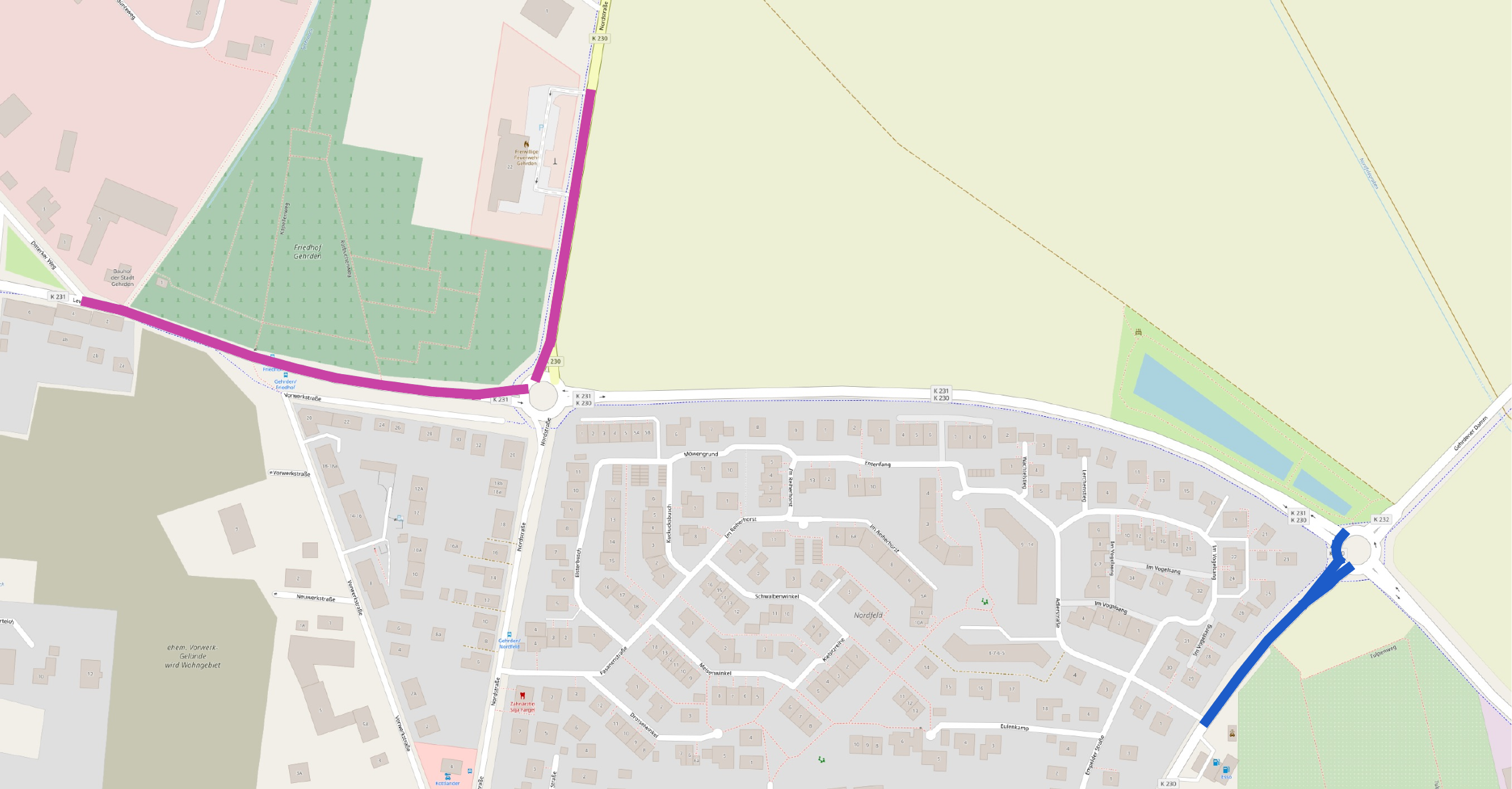}};
        \node[above right, rectangle, fill=white, opacity=0.6, text opacity=1] at (picture.south west) {\tiny \textcopyright OpenStreetMap contributors, ODbL};
    \end{tikzpicture}%
    
    \caption{Example of structural dependencies in a urban road network
    observed near Gehrden, Germany.}
    \label{fig:structIntro}
\end{figure}

We illustrate structural dependencies in urban road networks at the example of the area of Gehrden - a town in the district of Hanover, Germany -  in Figure \ref{fig:structIntro}. This figure illustrates two subgraphs of the road network (marked blue and purple). Both subgraphs represent the feeder roads to the main highway (B65) connecting Gehrden and the city of Hanover that constitutes the main commuting route for the Gehrden inhabitants. 
During a period with a high traffic load (e.g. during a rush hour), these subgraphs are typically simultaneously congested due to the network topology.
Thus, we consider such subgraphs to be structurally dependent.

The existing literature on RC mainly identifies temporal patterns \cite{2018mdpirecurrent, AN2016515}.
However, we observe a lack of methods that investigate the influence of road network topology on RC.
Detection of such dependencies within complex road networks is not trivial, particularly due to the variety of the influence factors (e.g. planned special events, accidents, construction sites and extreme weather conditions) and their dynamic impact on the traffic flow concerning the spatial and temporal dimensions.
To the best of our knowledge, the task of the data-driven discovery of structural dependencies in urban road networks is new and has not been addressed in the literature.

This article presents \approach - a novel unsupervised data-driven spatio-temporal data mining algorithm to reveal structural dependencies in urban road networks.
\approach relies on the intuition that structural dependencies can manifest as correlations of congestion patterns.
In this article, we represent congestion patterns as subgraphs of the road network.
We aim to exclude RC patterns that are mainly induced by temporal factors such as rush hour patterns that mainly depend on the daytime.
To this end, we only consider temporal traffic load outlier that factor out daily patterns to construct the subgraphs.
We identify the subgraphs using spatial clustering of the connected network segments that indicate a high traffic load. 
We consider temporal correlations of subgraphs located in spatial proximity as indicators of structural dependencies in the underlying road network.

To assess the effectiveness of the proposed \approach algorithm, we conducted a case study. This case study utilises two real-world traffic datasets in the regions of Hanover and Brunswick, Germany. The study results illustrate that our method can accurately detect meaningful structural dependencies in urban road networks.

In summary, our contributions are as follows: 
(1) We introduce the new task of the data-driven discovery
of structural dependencies in road networks;
(2) We propose the novel \approach algorithm to detect structural dependencies using traffic flow data; and 
(3) We conduct a case study to assess the effectiveness of the proposed method.

This article extends our prior work \cite{10.1145/3347146.3359109} towards this direction. 
Compared to \cite{10.1145/3347146.3359109}, in this article, we provide a detailed explanation of the individual algorithmic steps of \approach and the run-time complexity analysis. Furthermore, we present an extensive evaluation, including the manual assessment of the identified topological dependencies and a detailed investigation of the algorithm’s parameter impact.
A demonstration that includes visualisation of the dependencies identified by \approach in an interactive traffic analytics dashboard is available at \cite{10.1145/3397536.3422344}.

%% file: 2_relatedwork.tex
\section{Related Work}
\label{sec:related}

This section discusses related work in the areas of congestion analysis and spatio-temporal data mining, along with related data sources.

\subsection{Congestion Analysis}
Existing literature on congestion patterns distinguishes between Recurrent Congestion (RC) and Non-Recurrent Congestion (NRC) events \cite{doi:10.3141/1867-08}.
Non-Recurrent Congestion is defined as congestion that occurs because of singular events such as accidents \cite{wu2016interpreting,Pan:2015:FSI:2830193.2830267, 9040415}, extreme weather conditions \cite{7396485, CHUNG2012167,8283291}, or large-scale public events \cite{tempelmeier:2019,DBLP:conf/itsc/KwoczekMN15}.
Existing research on NRC addressed a variety of problems such as delay estimation \cite{tempelmeier:2019,DBLP:conf/itsc/KwoczekMN15}, routing adaptation \cite{Pan:2015:FSI:2830193.2830267}, congestion prediction \cite{7396485,8283291}, and NRC detection and tracking \cite{ANBAROGLU201447,wu2016interpreting, CHUNG2012167}.
Methods for NRC detection include spatio-temporal clustering \cite{ANBAROGLU201447}, regression models \cite{7396485}, and random forests \cite{8283291}.

Recurrent congestion denotes the remaining congestion events.
Rush hour patterns constitute the most prominent examples of RC events \cite{LAFLAMME201731}. 
A large number of studies focus on the temporal analysis of RC.
One line of research addresses the prediction of RC where machine learning models such as neural networks \cite{7966128, ijgi9030152, 8916966} or Support Vector Machines \cite{8481486} currently constitute the state-of-the-art.
The closely related task of short-term traffic forecasting is well studied in the existing literature \cite{VLAHOGIANNI20143}.
Models for both RC prediction and short-term traffic forecasting typically utilise periodically reoccurring traffic patterns, e.g. rush hours and day of week patterns, to facilitate predictions.
Another line of research investigates the evolution of RC patterns. Current approaches typically analyse the propagation of RC within a spatial grid \cite{2018mdpirecurrent, AN2016515, 8943987} or a road network graph \cite{10.1145/3283207.3283213, 8534362, SAEEDMANESH2017962}.

While these studies mainly address the temporal aspects of RC, we observe a lack of research that investigate the influence of road network topology on RC. 
This article proposes an algorithm that filters out periodic traffic patterns and identifies mutual dependencies of subgraphs within the road network.

\subsection{Spatio-Temporal Data Mining}
Spatio-temporal data mining algorithms address the challenge of extracting information, e.g., frequent patterns or anomalies, from large sets of spatio-temporal data.
Atluri et al. provide an overview of approaches and problems in a recent survey \cite{10.1145/3161602}.

Previous data mining approaches for road network data often aim at identifying individual important roads or junctions within the road network.
\cite{DBLP:journals/epjds/GuoZFTZLLH19} introduced a data-driven approach to identify the importance of individual roads within the road network
by measuring the correlation of traffic load between a particular road and the whole road network.
Similar, \cite{8357949} discovers individual important intersections. The authors represent trips within the road network as a tripartite graph. They compute the importance of intersections with an iterative ranking algorithm.
In contrast, we consider the problem of identifying pairwise dependencies
in the road network.

Several studies consider outlier detection in road traffic data. In \cite{brunauer2018recognizing} anomalous traffic flow is detected by grouping road intersections via their traffic flow patterns and self-organising maps. \cite{li2009temporal} focuses on detecting outliers in the traffic load by sudden changes. \approach builds upon existing outlier detection methods and exploits outlier co-occurrences and mutual information to determine spatio-temporal dependencies.

\subsection{Data Sources}
Road traffic data is often collected from stationary sensors, GPS devices, or simulations.
Stationary sensors, such as induction loops permanently installed within roads, are traditional traffic data sources. 
Stationary sensors usually measure high quality and consistent traffic data but lack coverage of the road network, especially in urban environments.
Existing research has widely adopted the use of data from stationary sensors \cite{Pan:2015:FSI:2830193.2830267,CHUNG2012167,ANBAROGLU201447,7966128}. 
The increasing digitisation of urban traffic has led to a boost in real-world traffic data availability. In particular, floating car data (FCD) is usually collected from GPS devices. FCD enables detailed and realistic insights into the specific regions' actual traffic load.  
FCD has proven to be suitable data source for the analysis of both RC (\cite{AN2016515,8534362}) and NRC (\cite{ASAKURA2017330}).
Compared to data collected from stationary sensors, FCD typically covers a larger fraction of the road network but is less consistent. 
%
Simulation-based approaches (e.g. \cite{taylor2008critical, scott2006network})
utilise the features originating from the network topology and capacity 
and can reveal critical parts of road networks and the possible impact of incidents.
However, these methods are restricted by the approximations of the underlying models that can provide only rough estimates of the real traffic flow.
In this article, we rely on FCD, representing real-world traffic flow data and can provide insights into the topological dependencies that become apparent only under real traffic conditions.

%% file: 3_problem.tex
\section{Problem Statement and Formalisation}
\label{sec:problem}
\newcommand{\au}{\textit{affected}(u,t)}
\newcommand{\asg}{\textit{affected}(sg,t)}

In this article, we address the problem of identification of the  
structurally dependent subgraphs in a road network. 
We consider subgraphs to be structurally dependent if: 
\begin{enumerate}
    \item The subgraphs are located in spatial proximity.
    \item The subgraphs are typically simultaneously affected by Recurrent Congestion. 
    \item The road network topology causes the correlation of the congestion on these subgraphs.
\end{enumerate}
In the following, we formalise the key concepts adopted in the article. 
In this formalisation, we adopt and, where necessary, extend some of the concepts defined in our previous work \cite{10.1145/3347146.3359109}.

\textbf{Transportation graph.} 
We represent the road network as a directed multi-graph $TG:=(V,U)$, referred to as a \textit{transportation graph}. 
$U$ is a set of edges (i.e. road segments); $V$ is a set of nodes (i.e. junctions).
We refer to an edge of the transportation graph as a \textit{unit} $u\in U$. 

\textbf{Unit load.} We denote the traffic flow observed on a unit at a particular time point as \textit{unit load}. 
Formally, $ul(u, t)$ is the traffic load on the unit $u$ at the time point $t \in \mathcal{T}$, where $\mathcal{T}$ denotes the set of time points.

We measure the \textbf{unit load} $ul(u, t) \in [0,1]$ as the relative speed reduction at unit $u$ at time point $t$ with
respect to the speed limit $lim(u)$ of the corresponding edge of the transportation graph:
\[ul(u,t)= \frac{lim(u) - speed(u, t)}{lim(u)},\]
where $speed(u, t)$ represents the traffic speed on unit $u$ at time $t$.

\textbf{Affected unit}. We denote a unit that exhibits an abnormally high traffic load at a certain time point $t$ as an \textit{affected unit} at $t$.   
Formally, \au: $U \times \mathcal{T} \mapsto  \{\textit{True}, \textit{False}\}$ indicates whether unit $u$ is affected at time point $t$.

\textbf{Subgraph.} We refer to a subgraph of the transportation graph as \textit{subgraph} $sg:=(V', U')$ with $V' \subset V$, $U' \subset U$.

\textbf{Affected subgraph}. An \textit{affected subgraph} represents a subgraph that exhibits an abnormally high traffic load at a certain time point $t$ (e.g. a congested highway section). Formally  $\asg:SG \times \mathcal{T}\mapsto \{\textit{True}, \textit{False}\} $ indicates whether a subgraph $sg$ is affected at time point $t$, where $SG=\{\mathcal{P}(V) \times \mathcal{P}(U)\}$ denotes the set of possible subgraphs, and $\mathcal{P}(\cdot)$ denotes the power set.

We consider a subgraph $sg$ containing the units $sg.U$ to be affected at time point $t$ if at least one of its units $u \in sg.U$ is affected at $t$:
\[
\asg =
\begin{cases}
    True, & \exists u \in sg.U: \au \\
    False,              & \text{otherwise}.
\end{cases} 
\]

%% file: 4_approach.tex
\section{Approach}
\label{sec:approach}

In this article, we aim to determine structurally 
dependent subgraphs in a road network.

\begin{figure}
    \includegraphics[width=0.95\textwidth]{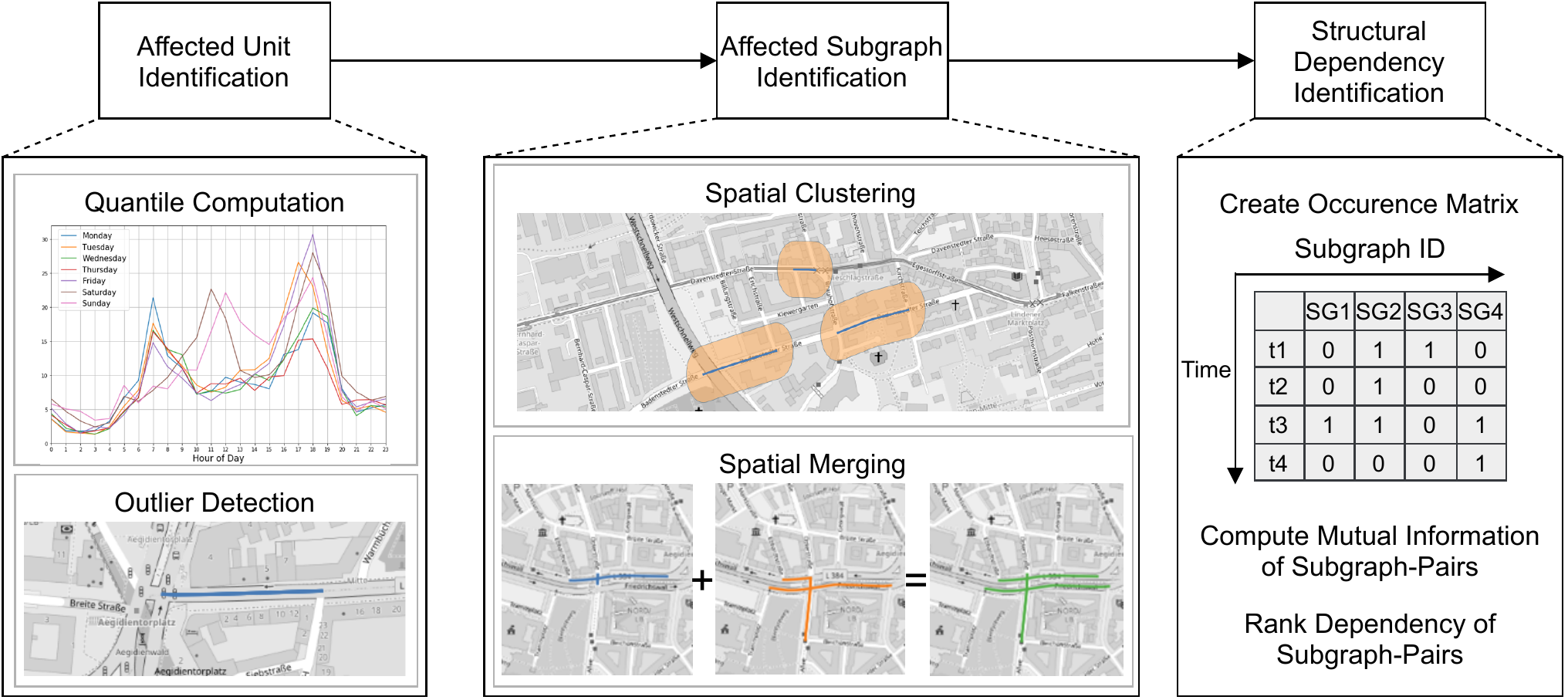}
    \caption{\approach pipeline overview. Map images: \textcopyright OpenStreetMap contributors, ODbL.}
    \label{fig:pipeline}
\end{figure}

Our \approach approach consists of the following main steps illustrated in Figure \ref{fig:pipeline}:
(i) We identify affected units of the transportation graph using traffic data. 
(ii) We develop algorithms to identify affected subgraphs of the transportation graph.
(iii) We develop an algorithm to identify structural spatio-temporal dependencies of the identified subgraphs.

\subsection{Identification of Affected Units}
\label{sec:affected-units}

This step aims to identify affected units, i.e. the units that exhibit an exceptionally high load at any single time point.
Factors influencing road traffic include reoccurring temporal factors (such as, e.g. rush hours) and spatial factors such as road network topology.
As we aim to identify Recurrent Congestion that result from the road network topology, it is desirable to exclude reoccurring temporal factors from our analysis.
To factor out exclusively temporal congestion patterns, we identify affected units as temporal outliers by employing the \emph{interquartile range} (IQR) rule \cite{iqrbook}.
As the weekday and the day time strongly influence the traffic load, we identify units that exhibit an exceptionally high load compared to the average traffic load on these units at the same weekday and day time.
More formally, we consider $u$ to be affected at time $t$, if the following condition holds: 
\[
\textit{affected}(u,t) =
\begin{cases}
    True, &  if~ul(u, t) > Q_3(u, t) + 1.5 \cdot IQR\\
    False,              & \text{otherwise},
\end{cases} 
\]
\noindent where $Q_n(u, t)$ denotes the $n^{th}$ quartile of the unit load on unit $u$ regarding the weekday and day time and $IQR=Q_3(u, t) - Q_1(u, t)$ denotes the interquartile range. 

We precompute the upper bound of the unit load $Q_3(u,t)+1.5 \cdot IQR$ for each unit, weekday and day time.
We consider each unit $u$ that exhibits a higher unit load $ul(u,t)$ than the corresponding upper bound to be affected at time point $t$.

\subsection{Identification of Affected Subgraphs}
\label{sec:clustering}

This step aims to identify topologically continuous, disjunctive subgraphs
of the transportation graph, homogeneous concerning their unit load at a given point in time. 
We approach this goal by conducting spatial clustering of the transportation graph's affected units. In this step, the clustering is performed independently at each point in time.
To ensure the spatial continuity of the resulting subgraphs, the clustering of affected units follows the basic region growing principle \cite{Bins:96}. 

We conduct the clustering as follows. First, we put one seed point on each affected unit. 
During the next steps, the regions are expanded by merging neighboured regions until there is no further change. The neighbourhood of two regions is determined by evaluating the distance between their closest edges in the transportation graph. 
This distance is measured as the edge count $d_{u}$ within the shortest path between the regions.

As the data can potentially be incomplete or contain measurement errors, we allow for certain tolerance while determining the neighbourhood. To this extent, we introduce the threshold $d_{u, max}$ to bridge gaps of a predefined size between two regions. Thus, two regions are considered as neighboured and are merged by the algorithm if the condition $ d_{u} \leq d_{u, max}$
holds. 
As a result, the units affected at time point $t$ are clustered into a set of $n$ clusters $C_t = \{c_0, \ldots, c_n \}$. 

Note that this clustering approach utilises the transportation graph's underlying graph structure. Thus, there is a unique mapping between each cluster and the corresponding subgraph of $TG$. 
In the following, while referring to the clustering results, we use the terms cluster and affected subgraph interchangeably. 
\par

\subsection{Spatial Merging of Affected Subgraphs}
\label{sec:merging}


The affected subgraphs identified in Section \ref{sec:clustering} are spatially disjoint with respect to the specific time points. Intuitively, when considering traffic load on the transportation graph over a longer time, we can observe spatial variations of the affected subgraphs, e.g. due to the congestion propagation along the graph. Furthermore, affected subgraphs (and their variations) 
can reoccur at different points in time. To capture these patterns, we conduct a merging of spatially overlapping affected subgraphs that occur at different time points.

Algorithm \ref{alg:mergeSubrahps} presents an incremental greedy approach to merge spatially overlapping affected subgraphs. The algorithm consist of a main loop (line 6-24) where the individual steps include 
candidate generation (line 9-11), 
similarity computation (line 12-14) and 
merging (line 15-24).
For the \emph{candidate generation}, we consider all subgraph pairs that share at least one unit as candidates (line 13).
Here, $[\mathcal{P}(\cdot)]^2$ denotes the subset of the power set with elements of cardinality 2.

\emph{Similarity computation} is performed for all candidates (i.e. subgraph pairs) by computing the similarity function $\texttt{similarity}:SG \times SG \mapsto [0,1]$ as follows (line 14):
\begin{gather*}
\texttt{similarity}(sg_1, sg_2) =
\begin{cases}
    1    & if ~sg_1 \subset sg_2 \lor sg_2 \subset sg_1,  \\
    \frac{|sg_1 \cap sg_2|}{|sg_1 \cup sg_2|},  & \text{otherwise.}
\end{cases} 
\end{gather*}%
Based on the definition of the affected subgraph, 
we consider the subgraph pairs in which one subgraph  entirely contains the other 
subgraph as a special case that has the maximum similarity of 1. 
Otherwise, the similarity is computed as \emph{Jaccard similarity} that measures to which extent the subgraph units overlap.

Finally, the \emph{merging} step aggregates the subgraph pairs with a similarity score above the threshold $th_{sim}$ (line 19-23).
The pairs with the highest similarity are merged first (line 16). 
Here the function \texttt{ordered}() orders the subgraph pairs in the descending similarity order. 
As merging affects the similarity computation, in each iteration of the algorithm, any subgraph can be merged only once (line 17-18).
%

The run time complexity of Algorithm \ref{alg:mergeSubrahps}  arises from combination of the main while loop in line 7 ($ \landau(|C|)$) and the iteration over the ordered set of subgraph pairs in line 16 ($\landau(|C|^2 \cdot \log(|C|^2)) = \landau(|C|^2 \cdot \log(|C|))$) resulting in a total complexity of $\landau(|C|^3 \cdot \log(|C|))$.

To facilitate an efficient candidate generation, we maintain a hashmap (\textit{commonUnits}) that provides a mapping from a single unit $u$ to all subgraphs that contain this unit $u$ (line 2-5).
The computation of all subgraph combinations 
would require quadratic time ($\mathcal{O}(|C|^2)$) in each iteration.
In contrast, the hashmap is computed once ($\mathcal{O}(|C|\cdot|U|)$) and is then updated iteratively during the algorithm according to the performed merging 
using the function \texttt{update}() (line 23).
The loops in line 10 and line 13 can be executed in parallel for further optimisation.

\begin{algorithm}
    \input{alg_Merge.tex}
\caption{Merge Subgraphs}
\label{alg:mergeSubrahps}
\end{algorithm}

\subsection{Identification of Spatio-Temporal Dependencies}
\label{sec:dependency-analysis}

In this step, we aim to identify dependent subgraphs of the transportation graph, i.e. the 
subgraphs that are typically simultaneously affected and are located in spatial proximity.
To this extent, we consider the subgraphs identified and merged in Section \ref{sec:merging}.
These subgraphs represent topologically connected subgraphs
of the transportation graph, including their spatial variations, that have been affected at some point(s) in time. In this step, we bring the temporal dimension into consideration and aim to identify the pairs of these subgraphs that are typically simultaneously affected.

Algorithm \ref{alg:dependencies} presents the method to identify such subgraph pairs, where the individual steps include candidate generation (line 8-15),
score computation (line 16-19) and sorting (line 20).

First, an occurrence matrix $occ[][]$ including the subgraphs and the time points is computed (line 1-7), where the columns correspond to the subgraphs and the rows to the time points.
If a subgraph is affected at time point $t$, then the corresponding cell is set to 1, otherwise to 0.
From the occurrence matrix, candidate subgraph pairs are generated (line 8-15). 
Each subgraph pair that is affected simultaneously in at least one time point is considered as a candidate pair.
For each candidate pair, we compute the subgraph dependency score. 
The intuition behind this score is to capture both the temporal co-occurrence and the spatial proximity of the subgraphs. Therefore, the score is computed as a combination of the mutual information and an inverse spatial distance metric:
\begin{gather*}
\texttt{dependency}(sg_1, sg_2) =
\begin{cases}
    0,  ~\textit{if} ~dist(sg_1, sg_2) \leq dist_{min} \\
    mi(sg_1,sg_2) \cdot \frac{1}{dist(sg_1, sg_2)}, ~otherwise.\\
\end{cases} 
\end{gather*}
Here, $dist(sg_1, sg_2)$ denotes the shortest geographic distance between two subgraphs. 
The threshold $dist_{min}$ specifies the minimum geographic distance for a subgraph pair to be considered dependent. $dist_{min}$ allows excluding trivial dependencies, such as adjacent subgraphs.
The mutual information $mi(sg_1,sg_2)$ aims to assess the temporal co-occurrence of two subgraphs, computed as:
\begin{gather*}
mi(sg_1,sg_2)=\sum_{t_{1} \in \mathcal{T}_1}\sum_{t_{2} \in \mathcal{T}_2}p_{(t_{1},t_{2})} (t_{1},t_{2})log \left (
\frac{p_ {(t_{1},t_{2})}(t_{1},t_{2})}{p_{t_1}(t_1)p_{t_2}(t_2)} \right ),
\end{gather*}
where $\mathcal{T}_i = \{t \in \mathcal{T} | ~\textit{affected}(sg_i, t)\}$ denotes the set of time points in which the subgraph $sg_i$ is affected.
The spatial proximity is measured as the inverse distance, where $dist(sg_1, sg_2)$ denotes the shortest geographic distance between two subgraphs. 
%
%
Finally, the subgraph pairs are ordered in the descending order of their dependency scores (line 20, \texttt{ordered}()).

The run time complexity of Algorithm \ref{alg:dependencies} results from the identification of candidates ($\landau(|SG|^2 \cdot \mathcal{T})$) in line 9 and line 12 as well as the sorting of subgraph pairs by score ($\landau(|SG|^2 \cdot \log(|SG|))$) in line 20.
Therefore, the overall complexity is bounded by $\landau(|SG|^2 \cdot (\mathcal{T} + \log(SG)))$.
Finally, the for loop in line 17 can be executed in parallel.

\begin{algorithm}
\input{alg_dependencies.tex}
\caption{Determine Spatio-Temporal Subgraph Dependencies}
\label{alg:dependencies}
\end{algorithm}

We provide an open-source implementation of the \approach algorithms under the MIT-license.\footnote{\url{https://github.com/Data4UrbanMobility/st-discovery}}

%% file: alg_Merge.tex
\begin{flushleft}
	\begin{tabular}{lll}

 	\texttt{Input}: & $C$: & Set of subgraphs  \\
 	\texttt{Output}: & $SG$: & Set of merged subgraphs   
	\end{tabular}
\end{flushleft}
\begin{algorithmic}[1]

    \STATE $SG \leftarrow C$

    \STATE $commonUnits \leftarrow []$
    
    \FORALL{$sg \in SG$}
        \FORALL{$u \in sg.U$}
            \STATE $commonUnits[u] \leftarrow commonUnits[u] \cup sg$
        \ENDFOR
        
    \ENDFOR

    \STATE changed $\leftarrow$ \texttt{True}
    
    \WHILE{changed}
        \STATE changed $\leftarrow$ \texttt{False}
        
        \COMMENT{Generate candidates}
        \STATE candidates $\leftarrow \emptyset$

        \FORALL{$u \in commonUnits$}
            \STATE candidates $\leftarrow$ candidates $\cup  [\mathcal{P}(commonUnits[u])]^2$
        \ENDFOR
        
        \COMMENT{Compute similarities}
        \STATE $s[] \leftarrow \emptyset$
        \FORALL{$(sg_1, sg_2) \in$ candidates}
            \STATE $s[(sg_1, sg_2)] \leftarrow \texttt{similarity}(sg_1, sg_2)$
            
        \ENDFOR
        
        \COMMENT{Merge subgraphs}
        \STATE visited $\leftarrow \emptyset$
        
        \FORALL{$(sg_1, sg_2) \in \texttt{ordered}(s)$ }
            \IF{$sg_1 \in \text{visited} \lor sg_2 \in \text{visited}$}
                \STATE \texttt{continue}
            \ENDIF
            \IF{$s[(sg_1, sg_2)] \geq th_{sim}$}
                \STATE $sg_1 \leftarrow sg_1 \cup sg_2$
                \STATE $SG \leftarrow SG 	\setminus sg_2$
                \STATE visited $\leftarrow \text{visited} \cup \{sg_1, sg_2\}$
                \STATE \texttt{update}(commonUnits)
                \STATE changed $\leftarrow$ \texttt{True}
            \ENDIF
        \ENDFOR

    \ENDWHILE
    \STATE \textbf{return}  $SG$

\end{algorithmic}

%% file: alg_dependencies.tex
\begin{flushleft}
	\begin{tabular}{lll}
	\texttt{Input}: & $SG$: & Set of subgraphs\\
	                & $\mathcal{T}$:& Set of time points \\
 	\texttt{Output}: & $P_{dependent}$  & Set of pairs of subgraphs, \\
 	  & & ordered by dependency score
	\end{tabular}
\end{flushleft}
\begin{algorithmic}[1]


\STATE $occ[][] \leftarrow \emptyset$
\FORALL{$t \in \mathcal{T}$}
    \FORALL{$sg \in \mathcal{SG}$}
        \IF{$\exists u \in SG: iqr(u,t)$}
            \STATE $occ[t][sg] \leftarrow 1$
        \ELSE
            \STATE $occ[t][sg] \leftarrow 0$
        \ENDIF
    \ENDFOR
\ENDFOR

\COMMENT{Determine candidate pairs}
\STATE candidates $\leftarrow \emptyset$

\FORALL{$(sg_1, sg_2) \in [\mathcal{P}(SG)]^2$}

    \IF{$(sg_1, sg_2) \in$ candidates}
        \STATE \texttt{continue}
    \ENDIF

    \FORALL{$t \in \mathcal{T}$}
        
        \IF{$occ[t][sg_1]=1 \land occ[t][sg_2] = 1$}
            \STATE candidates $\leftarrow$ candidates $\cup \{(sg_1, sg_2)\}$
            \STATE \texttt{break}
        \ENDIF

    \ENDFOR
\ENDFOR

\COMMENT{Compute dependency}

\STATE $P_{dependent} \leftarrow []$
\FORALL{$(sg_1, sg_2) \in candidates$}
    \STATE  score $\leftarrow \texttt{dependency}(sg_1, sg_2, \mathcal{T})$
    \STATE $P_{dependent}[(sg_1, sg_2)] \leftarrow score$

\ENDFOR
\STATE \textbf{return} $\texttt{ordered}(P_{dependent})$
\end{algorithmic}

%% file: 5_datasets.tex
\section{Datasets}
\label{sec:dataset}

\subsection{OpenStreetMap}
\label{sec:dataset_osm}
OpenStreetMap (OSM)\footnote{\url{https://www.openstreetmap.org}} is a provider of publicly available map data. We make use of the OSM road network to form the transportation graph $TG$. 
OSM partitions roads in smaller road segments that correspond to the transportation graph units $TG$. In particular, we extract the road segments located within the cities of Hanover and Brunswick, Germany. 
Considering the OSM-taxonomy for road types, we restrict the transportation graph to the major roads, as reliable traffic information for smaller roads is rarely available.
In particular, we extract all roads that belong to one of the following classes:
\{primary, primary\_link, secondary, secondary\_link, tertiary, tertiary\_link, motorway,  motorway\_link, trunk, trunk\_link\}. 
The extracted transportation graphs contain approx. 23,000 units (Hanover) and 7,600 units (Brunswick) in total. For each unit $u \in TG$, available information regarding the speed limit $lim(u)$ as well as the road type is extracted from OSM.

\subsection{Traffic Dataset}
\label{sec:dataset_fcd}
The experiments conducted in this article employ a proprietary traffic dataset that contains aggregated floating car data. In particular, the dataset provides traffic speed records for each unit $u$ of the transportation graph $TG$. 
The dataset was collected by a company offering routing software. The dataset contains data contributions obtained from various sources, including the data collected from the users of the routing software and traffic data acquired from third-party data providers. 
Although detailed statistics of these contributions, such as the number or the types of the monitored vehicles, are not available to the authors, due to the variety of sources involved we do not expect any particular biases towards certain vehicle types or expense classes. 

\begin{table}
    \caption{Dataset statistics for Hanover and Brunswick.}
    \centering
    \begin{tabular}{lcc}
    \toprule
     & Hanover & Brunsiwck \\
     \midrule
    No. Units  & 23,125 & 7,678  \\
    No. Records & $195 \cdot 10^6 $ &  $43 \cdot 10^6 $  \\
    Avg. No. Records/Unit & 8,422.79  & 5,674.91 \\
    Time Span & Oct 2017 - Jan 2018 & Dec 2018 - Jan 2019 \\
    \bottomrule
    \end{tabular}
    \label{tab:datasets}
\end{table}

The traffic data records contain the average traffic speed on the individual transportation graph units at discrete time points, i.e. $speed(u, t)$, recorded every 15 minutes.
The average speed records are computed by the data provider through calculating the average traffic speed from the raw floating car data, averaged over all vehicles for which the data is available for the given unit and time interval. 
The data for the major road categories mentioned in Section \ref{sec:dataset_osm} is captured regularly within the dataset. 
Table \ref{tab:datasets} presents statistics about the number of available traffic data records for Hanover and Brunswick.
We believe that the available data is sufficient to capture typical congestion patterns.\par

%% file: 6.0_evaluation.tex
\section{Experiments and Discussion}
\label{sec:experiments}

The evaluation aims to assess the effectiveness of the proposed \approach approach and its applicability to the real-world datasets presented in Section \ref{sec:dataset}.
First, we present an assessment of the dependencies identified by \approach.
Second, we evaluate and discuss the results of each main step of \approach, i.e., the identification of affected units and subgraphs, and the merging subgraphs.

\input{6.1_dependencies}

\input{6.2_affected_units}
\input{6.3_merging}

%% file: 6.1_dependencies.tex
\subsection{Structural Dependencies}
\label{sec:evaluation-spatio-temporal-dependencies}

The task of the data-driven identification of structural dependencies between subgraphs of a road network is new, such that neither a baseline nor any gold standard exists. Therefore, to assess the quality of the identified structural dependencies, we conduct a manual evaluation. In this evaluation, we use \approach to generate a ranked list of top-k subgraph pairs with high dependency scores, while using different values of $th_{sim}$.
To exclude trivial dependencies, i.e. the subgraphs that are adjacent to each other, we 
set the threshold $dist_{min}=500$ meters. 
We set $d_{u,max}=1$ for Hanover and $d_{u,max}=2$ for Brunswick.

We manually assess the correctness of each of the top-k subgraph pairs with the highest dependency scores. 
We judge each pair to be correct if we can observe and explain a structural dependency, or incorrect otherwise. 
The article authors performed the evaluation, while we discussed the individual judgments to obtain consensus.
Finally, we calculate the \emph{precision@k} as the proportion of the results judged as correct within the top-k subgraph pairs. 

\begin{figure}
    \centering
    \begin{subfigure}{0.45\textwidth}
    \includegraphics[width=\textwidth]{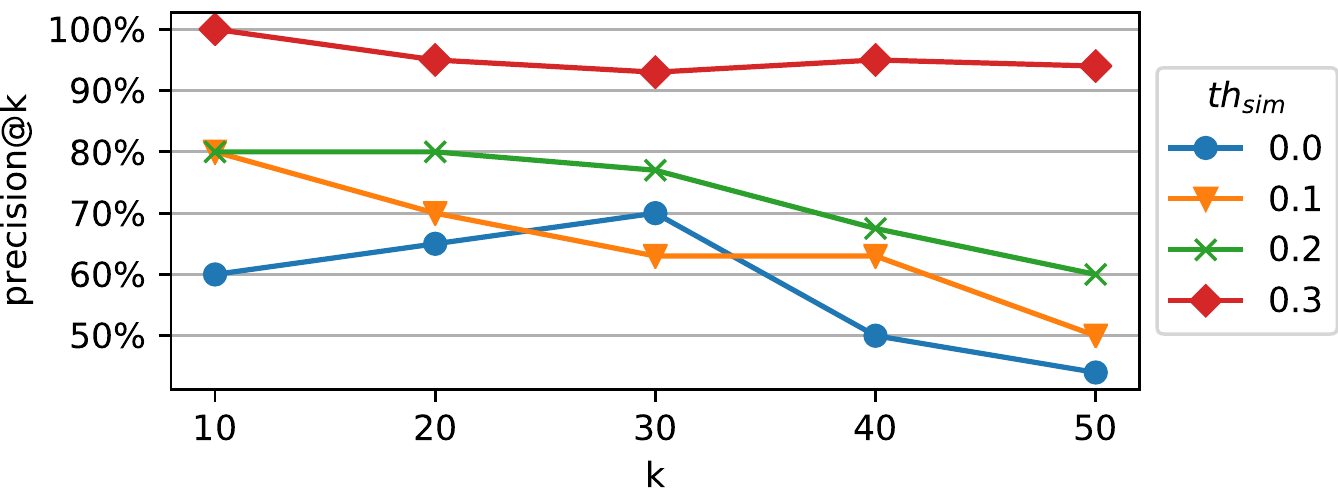}
    \caption{Hanover}
    \label{fig:subgraph_dependency_evaluation_han}

    \end{subfigure}
    \begin{subfigure}{0.45\textwidth}
    \includegraphics[width=\textwidth]{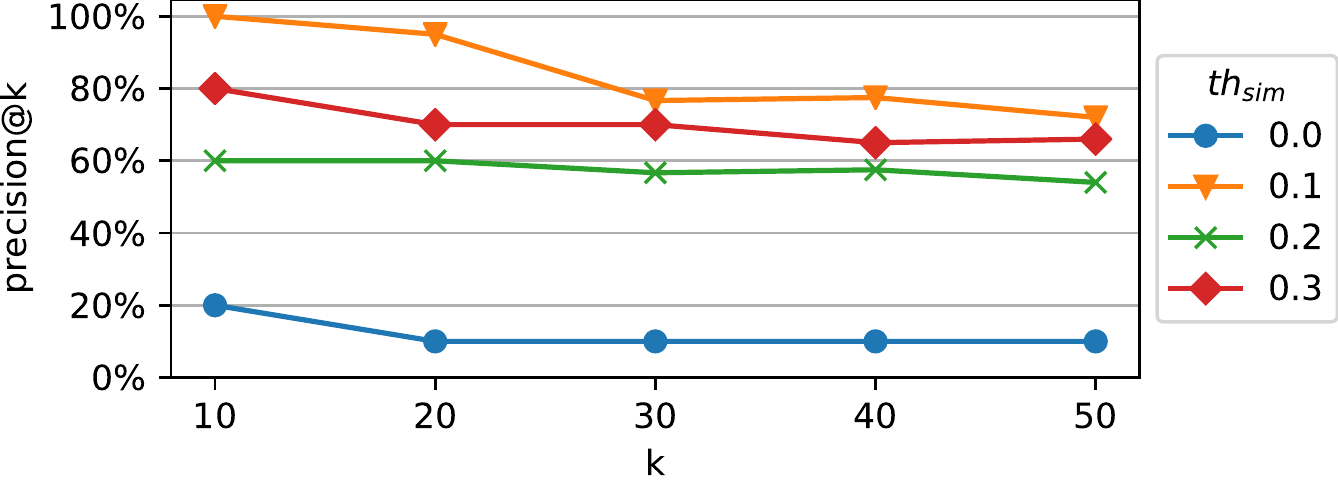}
    \caption{Brunswick}
    \label{fig:subgraph_dependency_evaluation_bra}
    \end{subfigure}

    \caption{Precision@k with respect to k and $th_{sim}$ of the identified structural subgraph dependencies.}
    \label{fig:subgraph_dependency_evaluation}
\end{figure}

Figure \ref{fig:subgraph_dependency_evaluation} presents the precision@k for $th_{sim} \in \{0, 0.1, 0.2, 0.3\}$ for Hanover (\ref{fig:subgraph_dependency_evaluation_han}) and Brunswick (\ref{fig:subgraph_dependency_evaluation_bra}).
For both datasets, the highest \emph{precision@k} is achieved at k=10 for all values of $th_{sim}$ except $th_{sim}=0$.
As the $k$ value increases (i.e., we consider more subgraph pairs with lower scores), the precision decreases in the majority of configurations. This behavior can be expected as the pairs with lower scores possess lower mutual information or are located at a higher geographic distance and are therefore less related. We conclude that the proposed score is well suited to quantify the dependency of affected subgraphs.

The best precision at k=10 is achieved by $th_{sim}=0.3$ (Hanover) and $th_{sim}=0.1$ (Brunswick). 
This indicates that the optimal value of $th_{sim}$ is dependent on the target road network.
In both cases, the worst performance is achieved at $th_{sim}=0$.
The graph partitioning at $th_{sim}=0$ is relatively coarse such that units that exhibit different dependencies can be aggregated into the same subgraph. 
Therefore, the achieved performance is lower than for the higher threshold values.

Note that the adopted evaluation method assesses the subgraph pairs' dependency but not the subgraphs' granularity.
Partitioning with higher threshold values (i.e. $th_{sim}=0.3$) leads to fine granular subgraphs. In this case, the partitioning can lead to a split of subgraphs that exhibit the same structural dependencies into different subgraphs, which may potentially lead 
to the inclusion of redundant subgraph pairs in the top-k results.

Therefore, whereas the combination of lower values of $k$ and higher values of $th_{sim}=0.3$ 
leads to the highest precision@k values, it can also lead to some redundancy in the top-k results (i.e. subgraph pairs representing the same dependency at different levels of granularity).
After manual inspection of the result granularity in our dataset, we observe that values of $th_{sim} \in \{0.1, 0.2\}$ lead to good results, with a precision of 80\% (Hanover) and 100\% (Brunswick) at k=10, while avoiding redundant subgraphs in the top results.
In general, we recommend that the $th_{sim}$ threshold should be adjusted according to the specific dataset and the use case under consideration.

\begin{figure}
    \centering
    \begin{subfigure}[b]{0.49\textwidth}
        \includegraphics[trim=0 25 0 25,clip, width=\textwidth]{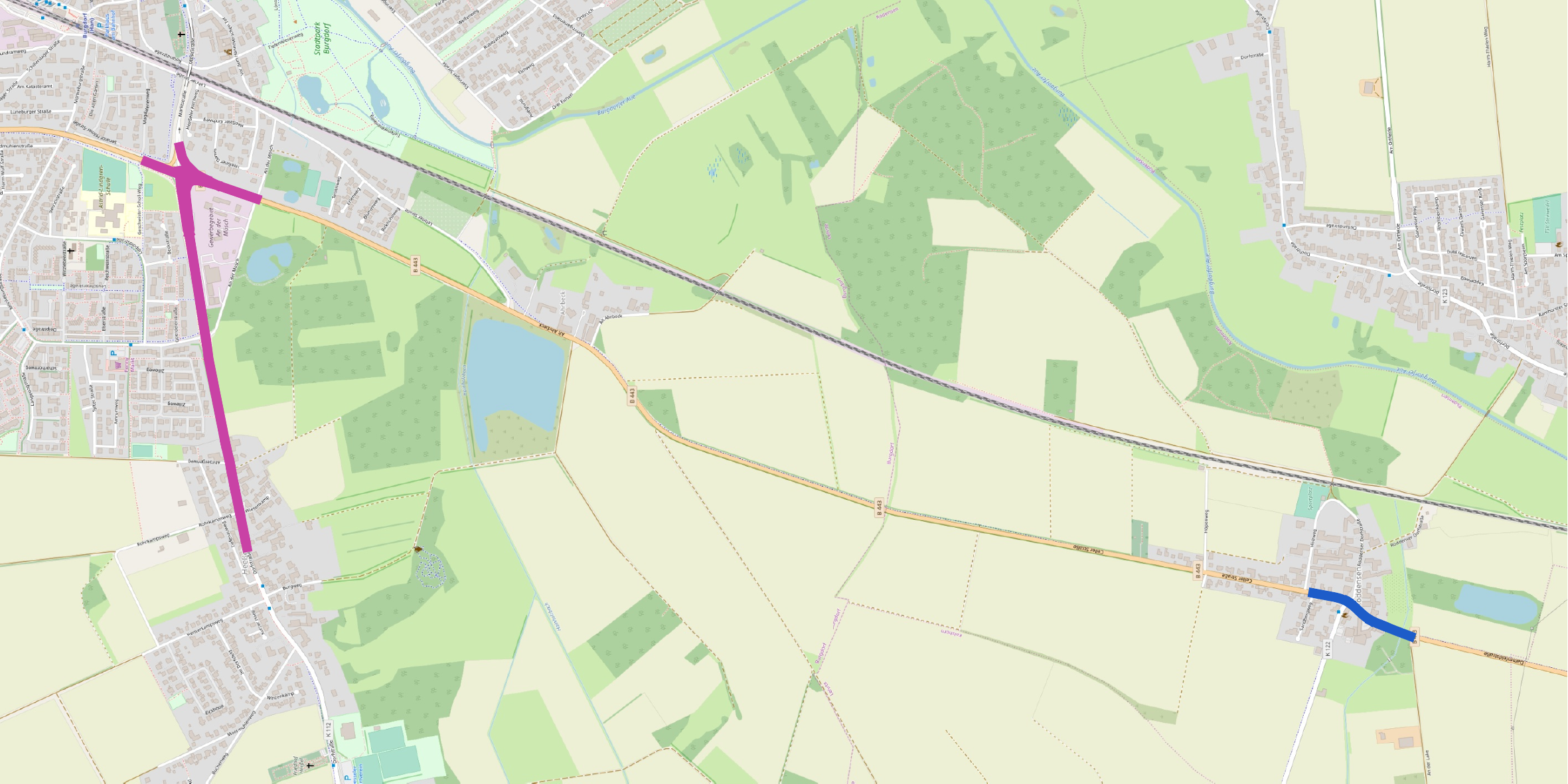}
        \caption{Junctions of a major rural street}
        \label{fig:evalDependExamples_junction}
    \end{subfigure}
    \begin{subfigure}[b]{0.49\textwidth}
        \includegraphics[trim=0 50 0 0,clip, width=\textwidth]{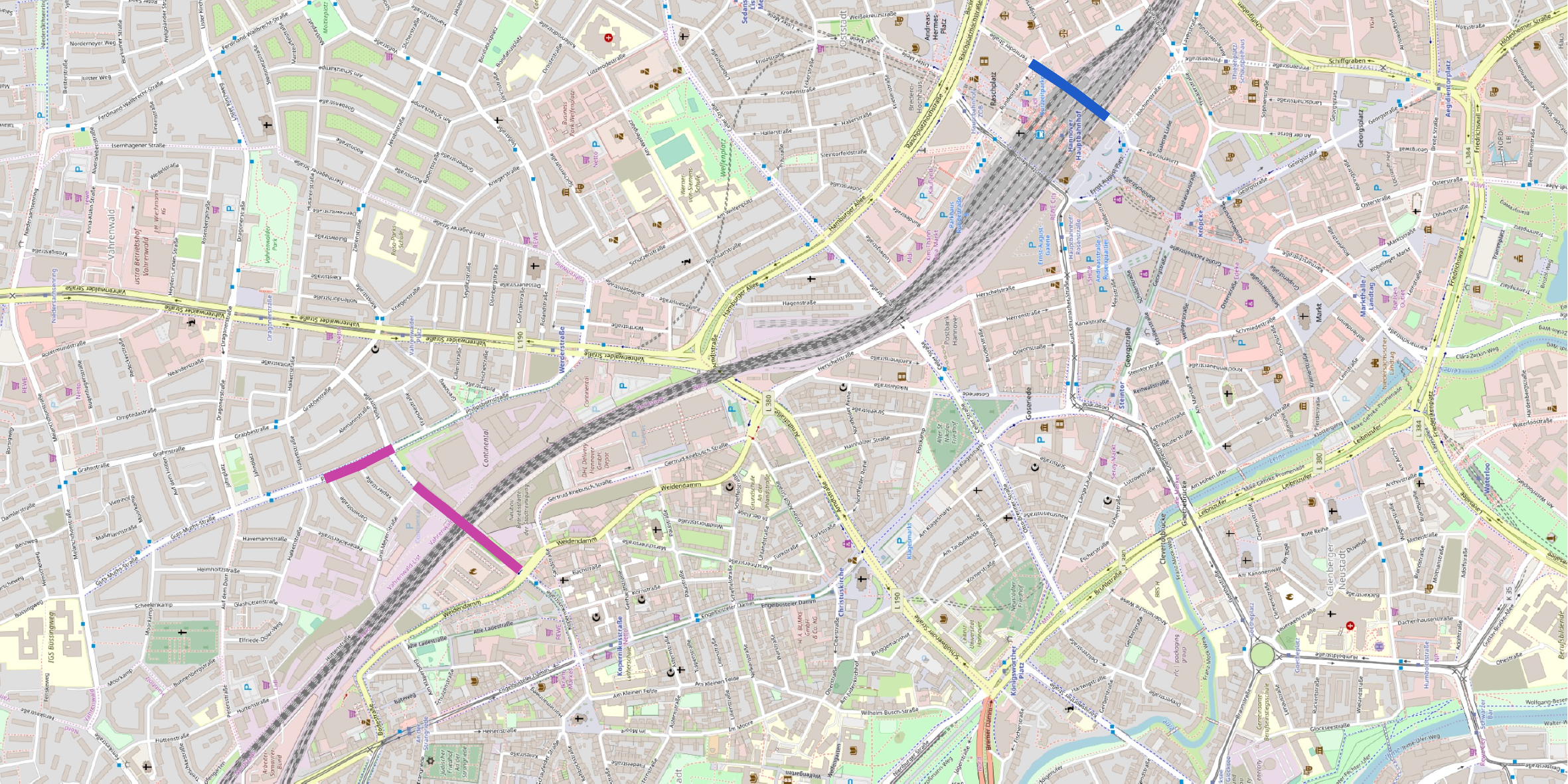}
        \caption{Streets crossing Hanover's central railway}
        \label{fig:evalDependExamples_railway}
    \end{subfigure}\vspace{0.2cm}
    \begin{subfigure}[b]{0.49\textwidth}
        \includegraphics[trim=0 40 0 10, clip, width=\textwidth]{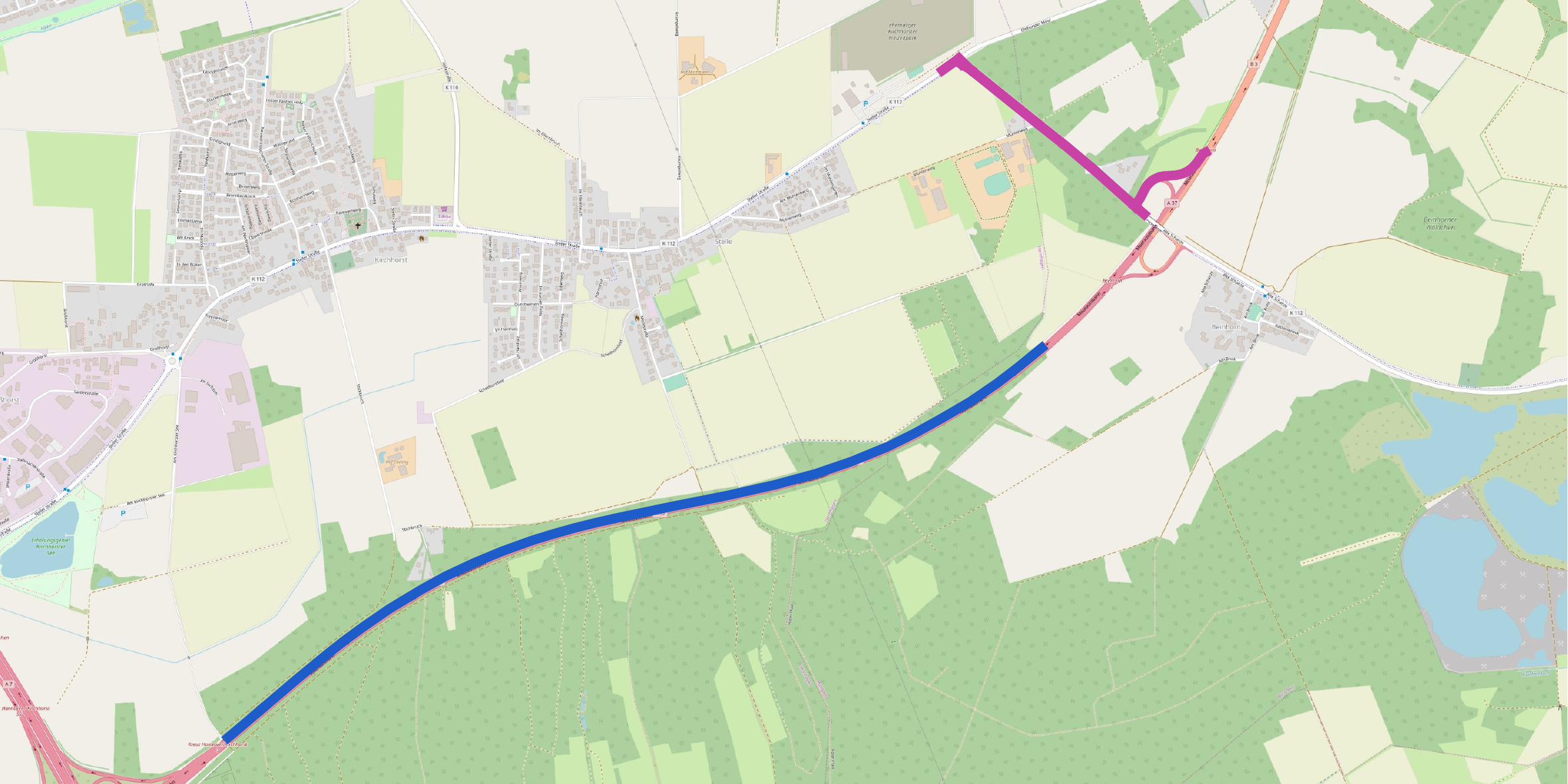}
        \caption{Highway and nearby exit}
        \label{fig:evalDependExamples_highway}
    \end{subfigure}
    \begin{subfigure}[b]{0.49\textwidth}
        \includegraphics[trim=0 50 0 00, clip,width=\textwidth]{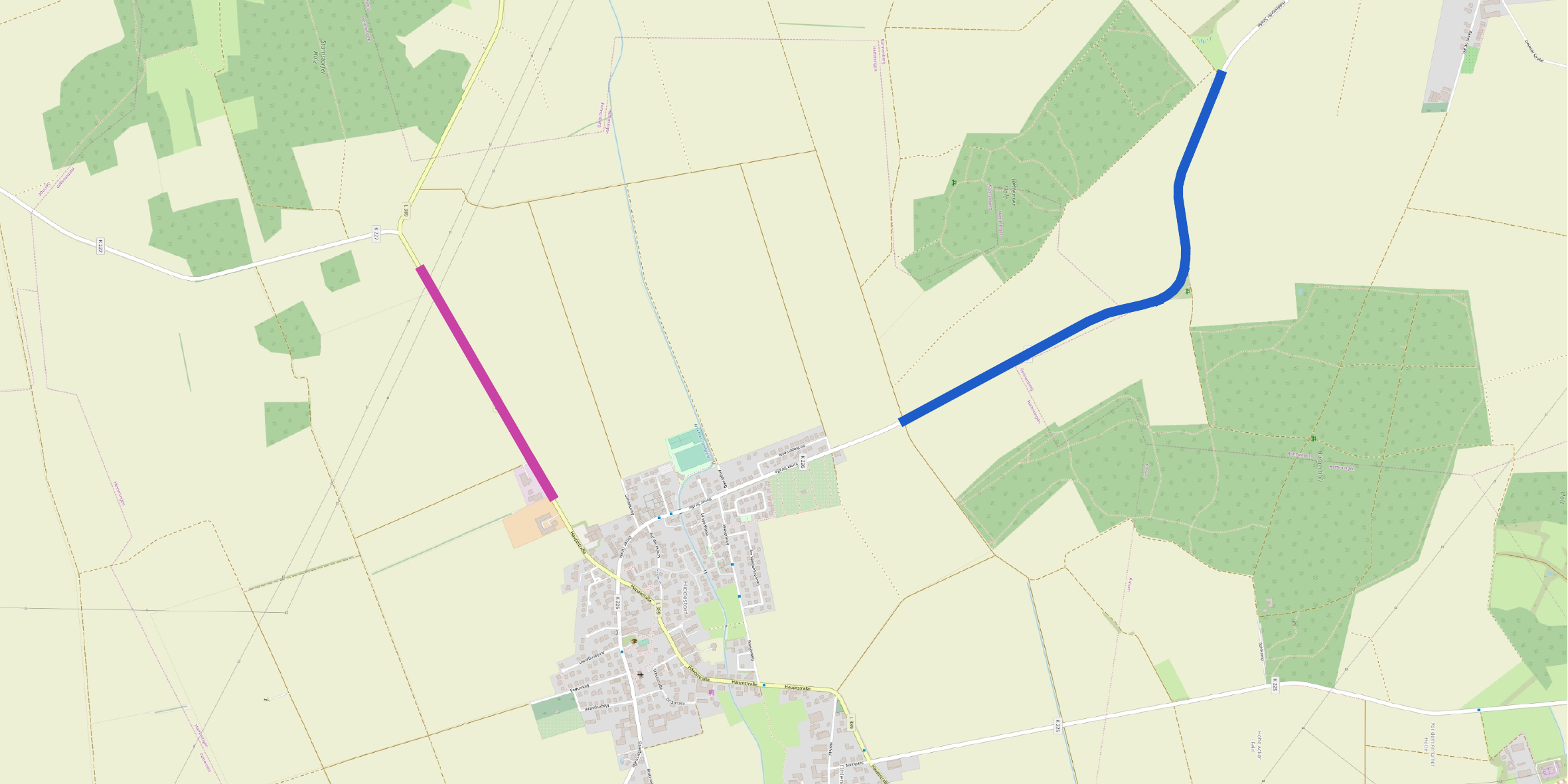}
        \caption{Roads leading to a town (purple) and leaving the town (blue)}
        \label{fig:evalDependExamples_town}
    \end{subfigure}
    \caption{Examples of the identified dependencies between subgraphs in Hanover. Dependent subgraphs are marked in blue and purple. Map images: \textcopyright OpenStreetMap contributors, ODbL.}
    \label{fig:evalDependExamples}
\end{figure}

To facilitate a better understanding of the determined dependencies, we discuss exemplary cases identified by \approach.
Figure \ref{fig:evalDependExamples} provides examples of the identified dependent subgraph pairs in Hanover, where the corresponding subgraphs in a pair are marked in blue and purple, correspondingly.
Figure \ref{fig:evalDependExamples_junction} shows two junctions of a major rural street.
The affected subgraph in the west includes the junction and its feeder roads, whereas only the junction is affected in the east.
This combination can be caused by the drivers who avoid the larger congestion in the west by accessing the street in the east, leading to increased traffic on both junctions.
Figure \ref{fig:evalDependExamples_railway} depicts two affected subgraphs that cross the central railway within the city of Hanover. The railway divides two city districts and needs to be crossed when travelling between these districts. Therefore the subgraphs represent alternative routes for trips from the north to the south and form a bottleneck for such trips.
Figure \ref{fig:evalDependExamples_highway} illustrates an affected highway (blue) and the last possible exit before that section (purple). If the highway is affected, the nearby exit and the consecutive roads, get affected as well. This is likely caused by drivers trying to exit the highway before entering the congested part, leading to an increased load in that region.
Figure \ref{fig:evalDependExamples_town} depicts a road leading to a town (purple) and another road leaving the town (blue). This indicates a large amount of traffic unnecessarily passing through the town to reach from the purple to the blue subgraph because of lack of alternative routes. In this case, building a new alternative road could prevent the town from being exposed to the high traffic load.

\begin{figure}[t]
    \centering
    \begin{subfigure}[b]{0.49\textwidth}
        \includegraphics[trim=0 20 0 20,clip, width=\textwidth]{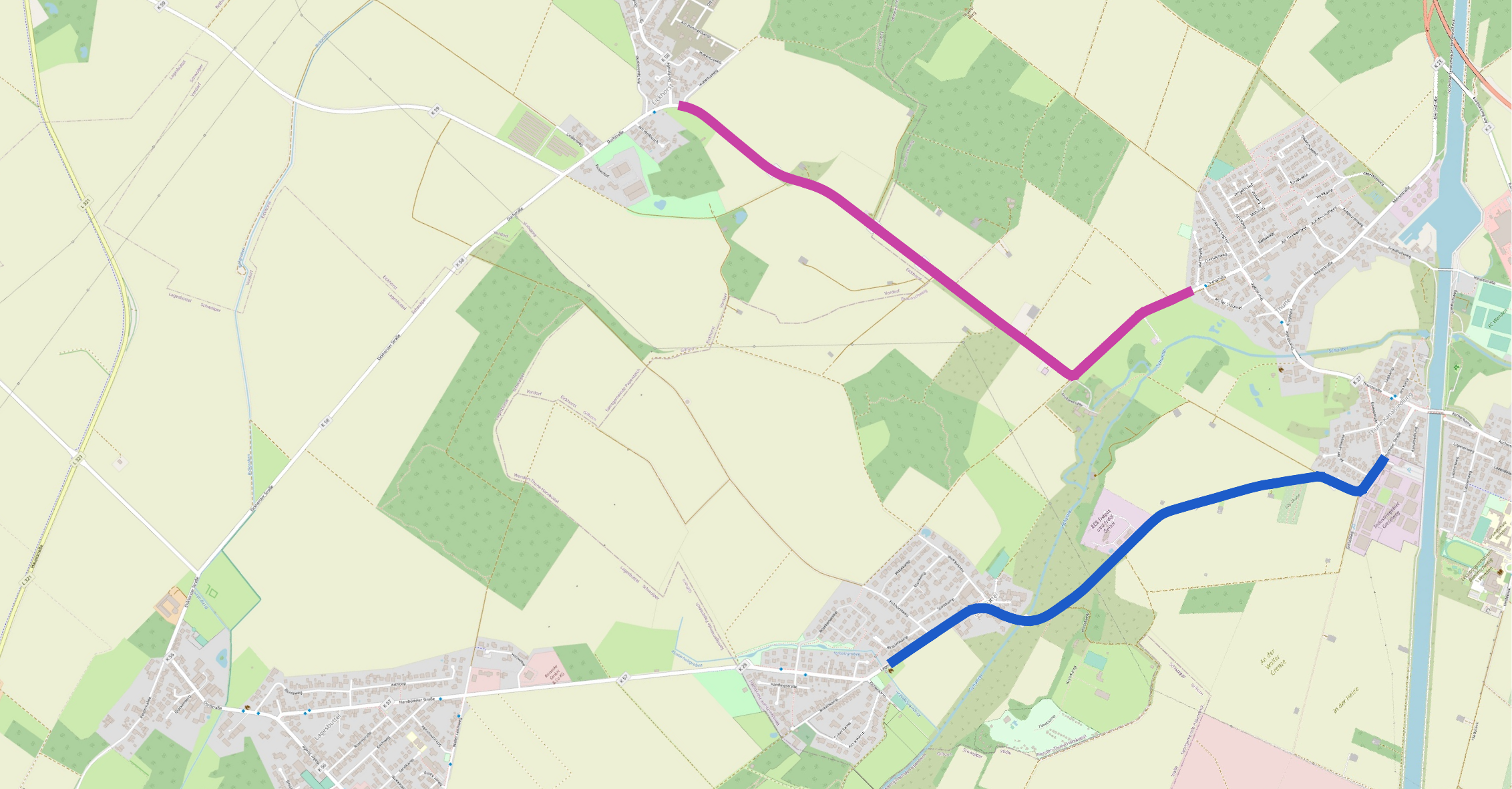}
        \caption{Near parallel rural roads}
        \label{fig:evalDependExamples_bra_parallel}
    \end{subfigure}
    \begin{subfigure}[b]{0.49\textwidth}
        \includegraphics[trim=0 10 0 30,clip, width=\textwidth]{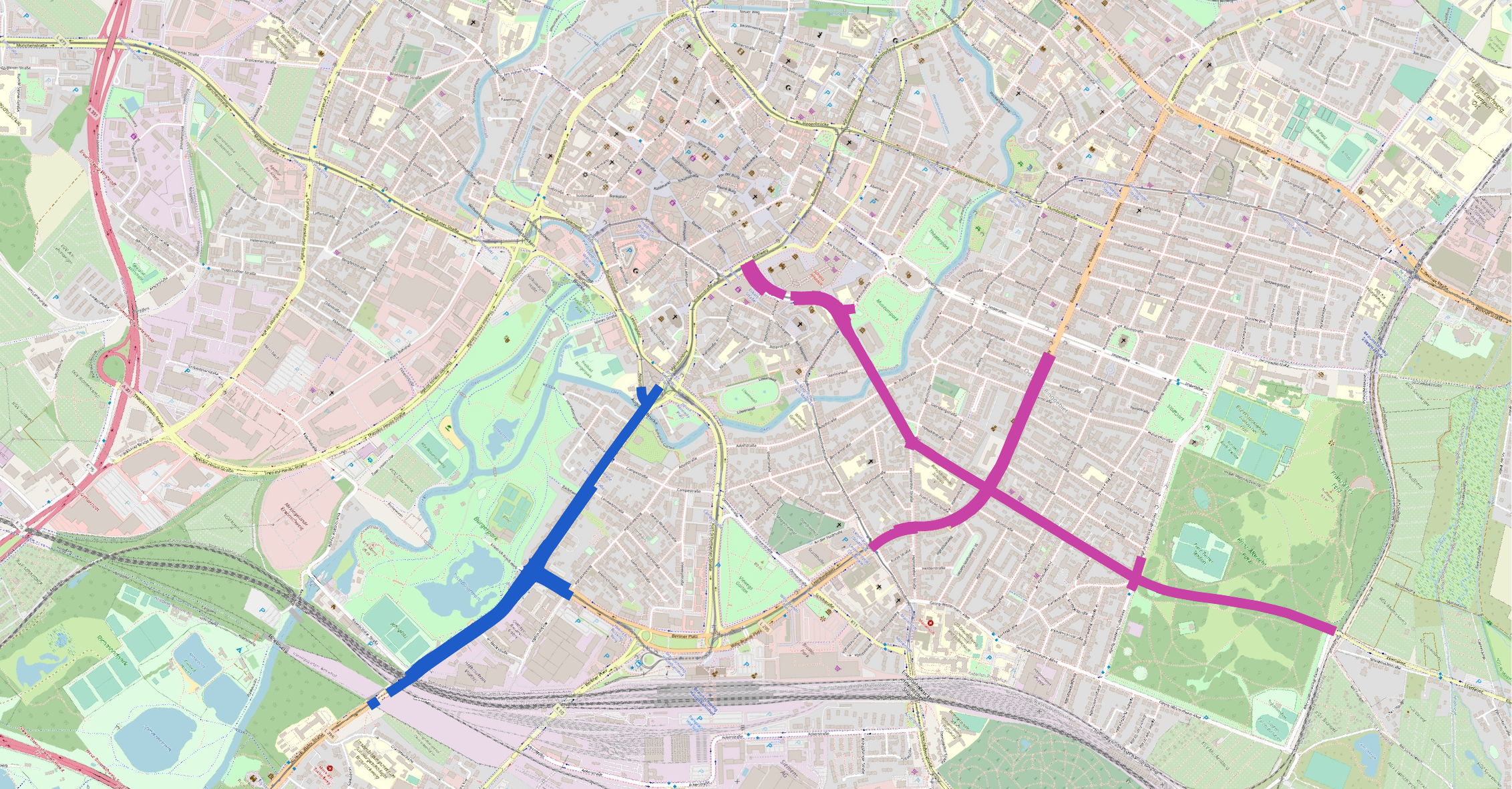}
        \caption{Streets sections leading to and from the city centre}
        \label{fig:evalDependExamples_bra_city_center}
    \end{subfigure}
    \caption{Examples of identified dependencies between subgraphs in Brunswick. Dependent subgraphs are marked in blue and purple. Map images: \textcopyright OpenStreetMap contributors, ODbL.}
    \label{fig:evalDependExamples_bra}
\end{figure}

Figure \ref{fig:evalDependExamples_bra} provides examples of identified dependent subgraph pairs in Brunswick.
Figure \ref{fig:evalDependExamples_bra_parallel} shows two near-parallel road sections that connect similar city districts. The roads constitute alternative routes for trips to the east or west. If one of the road segments faces congestion, the other road is likely to face increased load caused by drivers that want to avoid congestion.
Figure \ref{fig:evalDependExamples_bra_city_center} presents two dependent subgraphs near the city centre of Brunswick. The subgraphs are prominent options for leaving or entering the city centre. If the city centre is congested, the congestion is likely to propagate to the subgraphs as well. Furthermore, if one subgraph is congested, the other subgraph represents an alternative route for a similar trip.

%% file: 6.2_affected_units.tex
\subsection{Analysis of Affected Units and Subgraphs}
In this section, we analyse the distribution of the affected units and subgraphs identified by \approach in our datasets and the influence of the corresponding parameter.

\subsubsection{Distribution of Affected Units}
\label{sec:distribution-affected-units}

\begin{figure}
     \centering
    \begin{subfigure}{0.49\textwidth}
     \includegraphics[width=\textwidth]{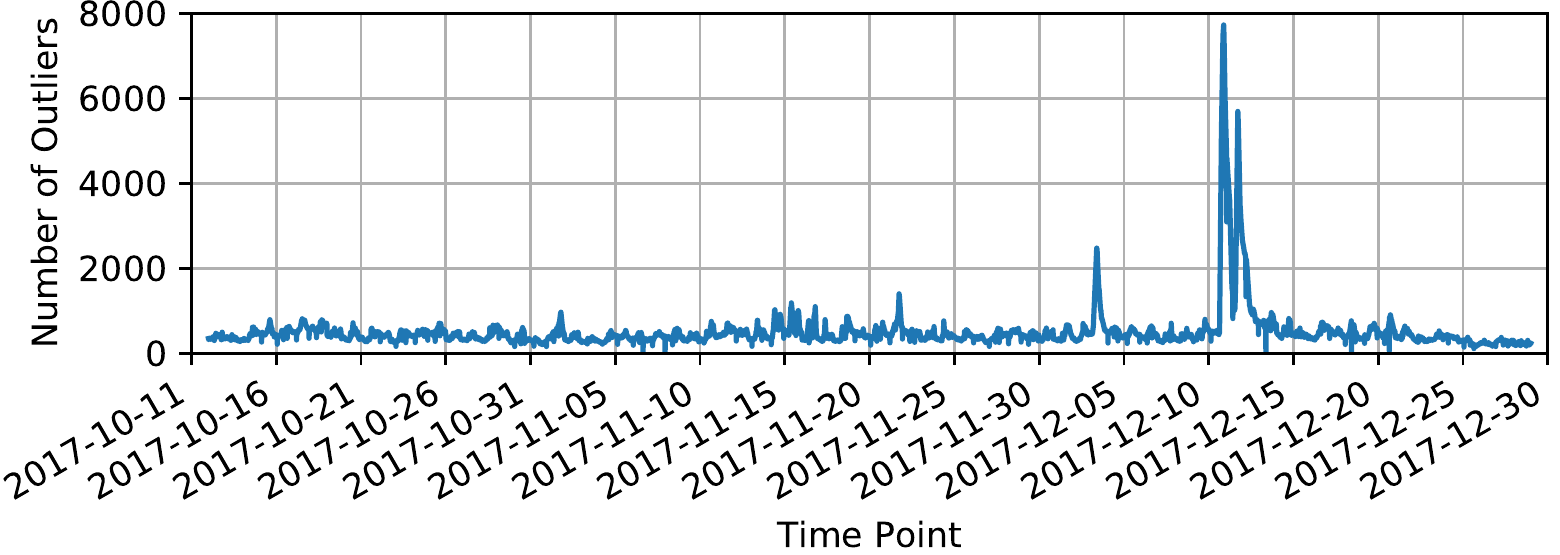}
     \caption{Hanover, between November 2017 and December 2017}
     \label{fig:outliersOverTime_Han}
    \end{subfigure}
    \begin{subfigure}{0.49\textwidth}
    \includegraphics[width=\textwidth]{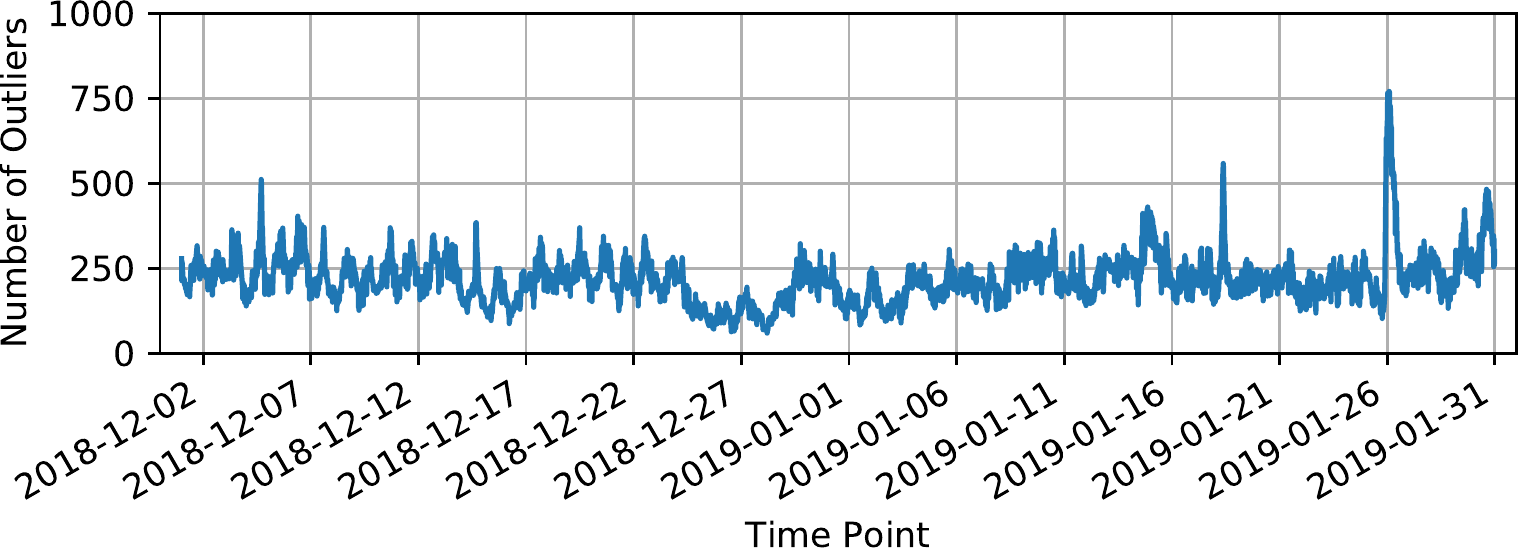}
    \caption{Brunswick, between December 2018 and January 2019}
    \label{fig:outliersOverTime_Bra}
    \end{subfigure}
     \caption{Number of affected units per time point.}
     \label{fig:outliersOverTime}
 \end{figure}

The analysis results of the distribution of affected units identified by the algorithm presented in Section \ref{sec:affected-units}, is shown in Figure \ref{fig:outliersOverTime}. 
Figure \ref{fig:outliersOverTime_Han} presents the number of affected units per time point between November and December 2017 in Hanover.
We observe that the number of affected units varies continuously, from 3 to 7724 in our dataset, with a median value of 409. 
Furthermore, we can observe several peaks (i.e. time points that exhibit an exceptionally high number of affected units). 
We observe the highest peaks between 2017-12-10 and 2017-12-15. 
Through a manual investigation (i.e. search for news articles related to this region and dates on Google), we found that heavy snowfall caused high and unusual delays on the whole road network during this period. This is reflected in a high number of affected units throughout the entire network.
Figure \ref{fig:outliersOverTime_Bra} presents the number of affected units per time point between December 2018 and January 2019 in Brunswick.
We observe a similar continuous variation of the number of affected units from 59 to 770 with a median value of 203.

Smaller peaks occur at several times, for instance at 2017-12-03 (Hanover) and 2019-01-26 (Brunswick), where the size of the peaks highly differs.
Given these observations, we believe that temporary peaks in the number of affected units can indicate occurrences of extraordinary incidents. 
Note that these observations are solely based on the temporal co-occurrences and do not provide any insights into the incidents' spatial characteristics.

\subsubsection{Distribution of Affected Subgraphs}
\label{sec:cluster-distribution}

\begin{figure}
    \hfill
    \begin{subfigure}{0.47\textwidth}
        \includegraphics[trim=4 4 4 4, clip, width=\textwidth]{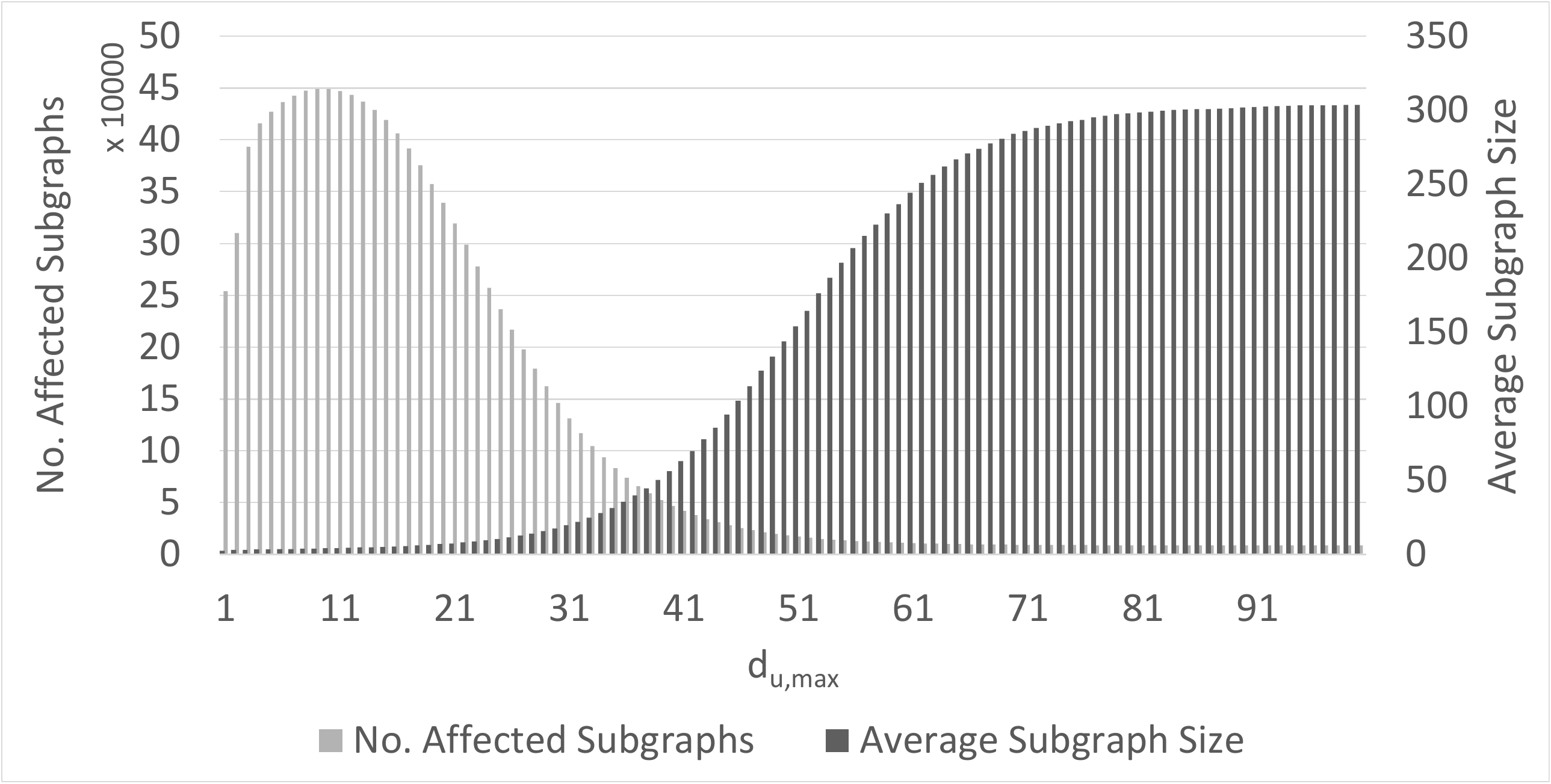}
        \caption{Hanover}
        \label{fig:evaluation_cluster_distribution_han}
    \end{subfigure}
    \hfill
    \begin{subfigure}{0.47\textwidth}
        \includegraphics[trim=4 4 4 4, clip, width=\textwidth]{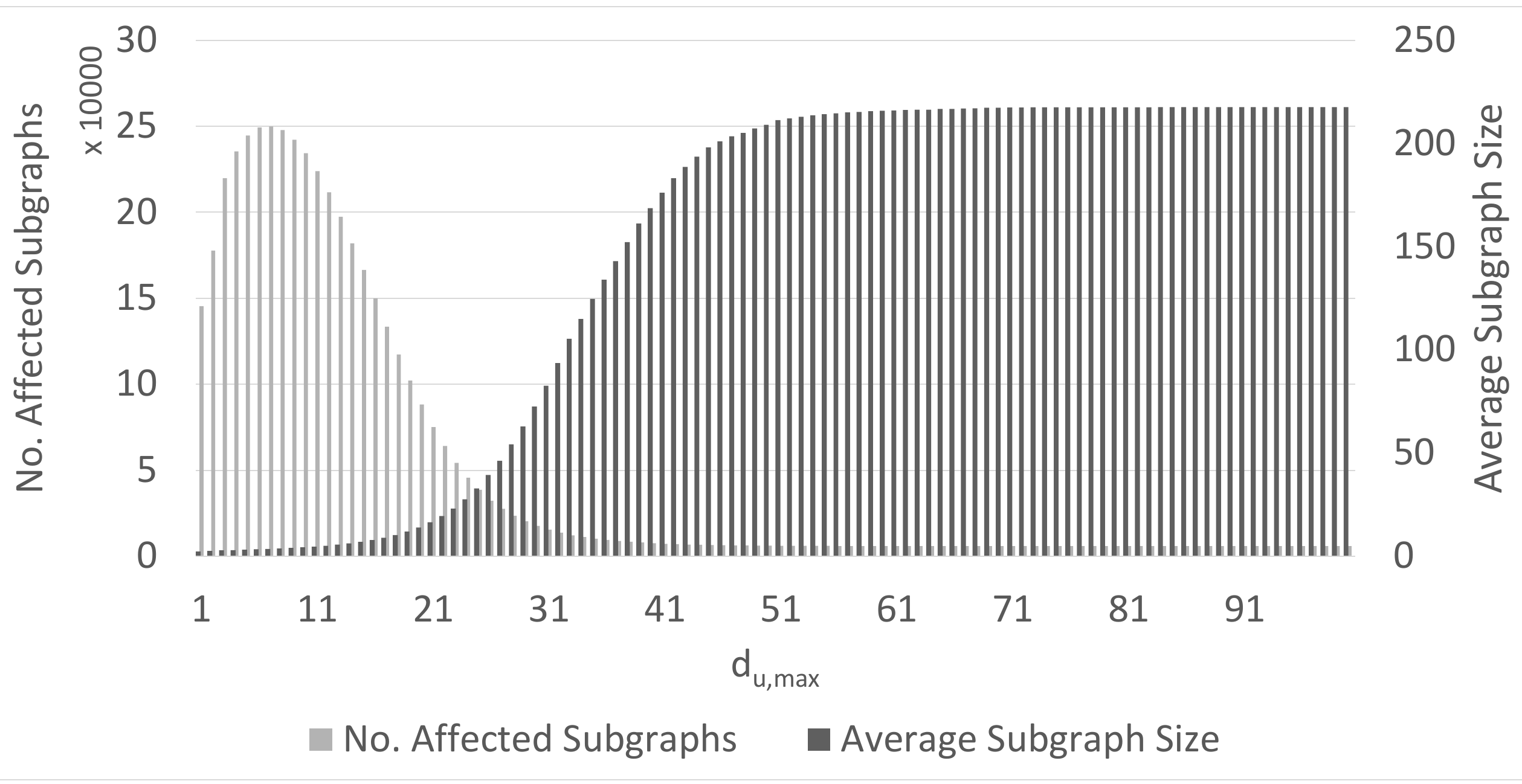}
        \caption{Brunswick}
            \label{fig:evaluation_cluster_distribution_bra}

    \end{subfigure}
    \hfill
    \caption{The number and the average size of the identified affected subgraphs show opposite trends dependent on the chosen threshold $d_{u, max}$. Subgraph size is measured as the number of edges.}
    \label{fig:evaluation_cluster_distribution}
\end{figure}

In this section, we analyse the distribution of the affected subgraphs 
identified using the algorithms presented in Section \ref{sec:clustering} and the influence of the relevant parameters.

During the identification of affected subgraphs using clustering of affected units presented in Section \ref{sec:clustering}, 
the threshold $d_{u, max}$ was introduced, which describes the distance tolerance in assigning affected units to a subgraph.
Therefore variations of the value of this threshold leads to a different number and size of unit clusters, as illustrated in Figure \ref{fig:evaluation_cluster_distribution_han} for Hanover and in Figure \ref{fig:evaluation_cluster_distribution_bra} for Brunswick.

With an increasing tolerance (i.e. the value of $d_{u, max}$) two trends can be observed in both road networks. While the average size of the resulting subgraphs increases, their number decreases. This is caused by the fact that more and more subgraphs are merged if the tolerance for determining a segment neighbourhood is increased.\par

Comparing the road networks with each other, we observe that the saturation of the subgraph size is reached at smaller values of $d_{u,max}$ for Brunswick (at 51) than for Hanover (at 71) since the road network of Brunswick (7678 units) is smaller than the road network of Hanover (23125 units).

Due to such opposite trends, the choice of a suitable value for $d_{u, max}$ strongly depends on the application scenario, the road network and the scale of the analysis. For instance, if a large region, e.g. a whole city or one of its districts, has to be analysed, tiny subgraphs consisting of just a few road segments are not that important. For this case, a higher threshold value should be chosen. In contrast, for a detailed inspection of smaller parts of the road network, e.g. specific roads or junctions, finer subgraphs are more critical. In this use case, to prevent merging of smaller subgraphs, a lower value of $d_{u, max}$ should be selected. 

\subsubsection{Temporal Persistence of Affected Subgraphs}
\label{sec:subgraph-persistence}

Besides the size of the identified affected subgraphs, we analyse their temporal persistence at the example of Hanover.
This means that the affected subgraphs identified at time point $i$ (including the typical variations of these subgraphs, e.g. due to the propagation of the traffic load along the transportation graph) have to be recognised in the subsequent time steps. For this purpose, we apply an algorithm based on the Hungarian method \cite{kuhn_hungarian_2012}. This algorithm aims to find an optimal assignment of clusters (i.e. affected subgraphs) in two consecutive time steps by minimising the assignment costs. In order to compute the assignment costs, the intersection of the affected units involved in each subgraph (cluster) in two subsequent time steps $i$ and $j$ is calculated as:
\[
 costs_{i,j} = ||c_i \cap c_j|| \ \ \ \forall c_i \in C_i, c_j \in C_j.
\]
Further, no tolerance would allow a subgraph to skip the subsequent time steps. This means the subgraph existence time will not be prolonged if this subgraph does not appear in a time step. Thus, if there is no assignment for a current subgraph in the following time step, this subgraph will 'die'. If there is a cluster of affected units located on the same road segments in the next but one time step, this cluster will be treated as a new affected subgraph, i.e. a new identifier will be assigned to this subgraph and a new existence time will be initiated. \par

\begin{figure}[t]
    \centering
    \includegraphics[trim=4 4 4 4, clip, width=0.7\textwidth]{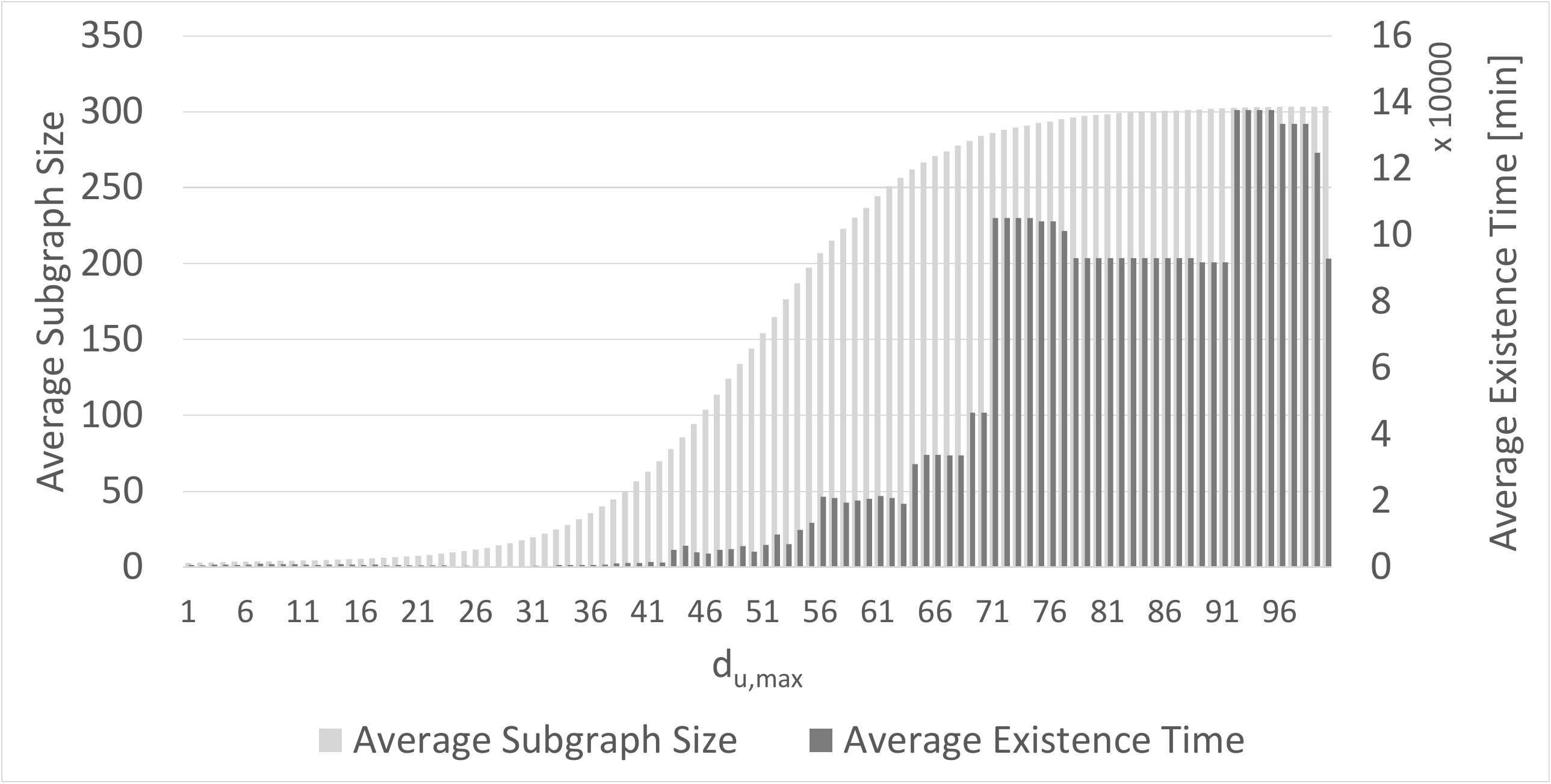}
    \caption{The existence time and the size of the affected subgraphs dependent on the value of the $d_{u, max}$ threshold.}
    \label{fig:evaluation_cluster_duration}
\end{figure}    

As we assume that the existence time of a cluster depends on its size, we evaluated the subgraph existence time in dependency of their size. 
As discussed in Section \ref{sec:cluster-distribution}, 
the size of the subgraphs depends on $d_{u, max}$. Figure \ref{fig:evaluation_cluster_duration} shows the average existence time of the clusters in relation to the average size and the chosen threshold $d_{u, max}$. The overall trend indicates that the affected subgraphs' existence time increases with their size. Thus, the choice of $d_{u, max}$ does not only influence the subgraph size but also its existence time.\par

%% file: 6.3_merging.tex
\subsection{Evaluation of Subgraph Merging}
\label{sec:evaluation-merging}

\begin{figure}
    \centering
    \includegraphics[width=0.6\textwidth]{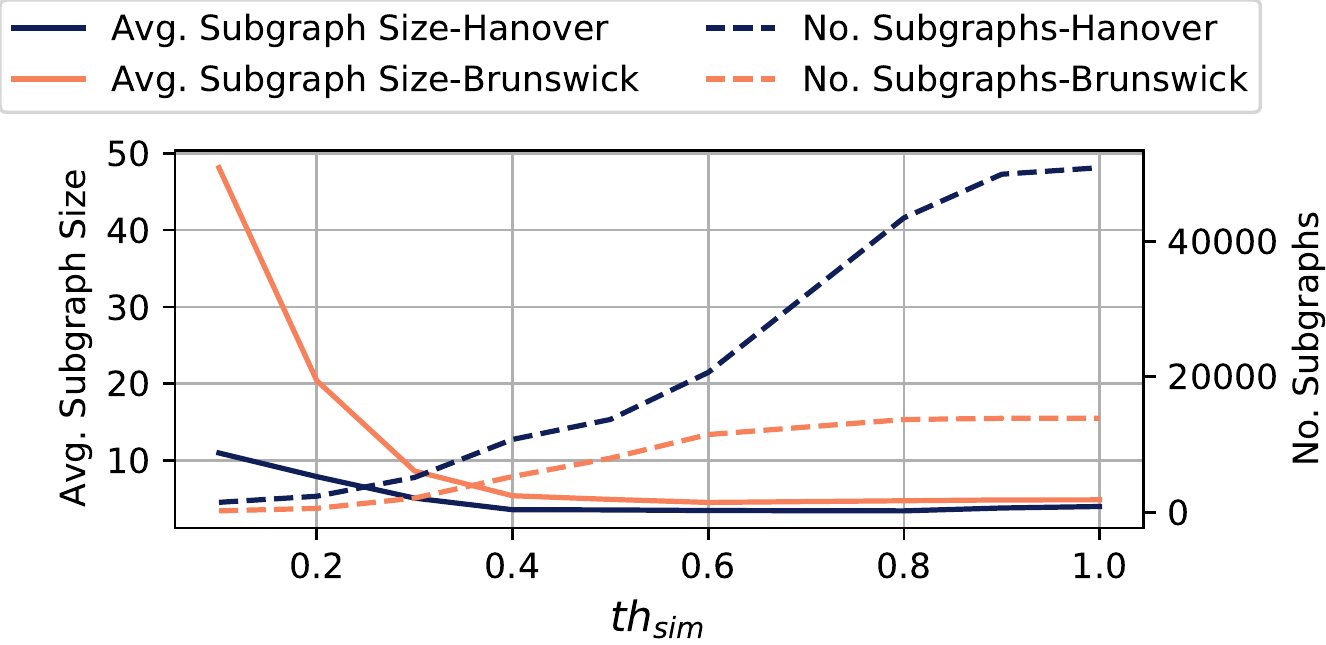}
    \caption{Influence of $th_{sim}$ in Algorithm \ref{alg:mergeSubrahps} on the number of subgraphs and an average subgraph size.}
    \label{fig:subgraphSize}
\end{figure}

In this section, we analyse the influence of the threshold $th_{sim}$ on the results 
of the subgraph merging conducted using Algorithm \ref{alg:mergeSubrahps}.
Figure \ref{fig:subgraphSize} opposes the number of subgraphs and the average size of subgraphs computed by Algorithm \ref{alg:mergeSubrahps} with respect to the similarity threshold $th_{sim}$ that specifies to which fraction two subgraphs need to overlap to be merged. Higher threshold values are more restrictive.

In general, as $th_{sim}$ increases, we observe a growing number of subgraphs, whereas the average size of these subgraphs decreases. This indicates that higher $th_{sim}$ values result in a finer granular division of the road network into smaller subgraphs.
The highest change of the average subgraph size can be observed for $th_{sim} \in [0, 0.4]$. For $th_{sim} > 0.4$ we only observe small changes of the average subgraph size. We conclude that values within $[0, 0.4]$ are particularly suited to calibrate the algorithm in the considered setup, whereas higher threshold values result in a larger number of subgraphs (up to 50k independent subgraphs in Hanover) and only weakly affect the subgraph size.

\input{subgraphs.tex}

We illustrate the influence of $th_{sim}$ at the example of a major junction in Hanover within our dataset. Figure \ref{fig:subgraphs} presents six different partitionings of the road network into subgraphs with respect to the values of $th_{sim} \in [0, 0.5]$, where the colour of the units represent their assignment to different subgraphs.
In general, the subgraphs become finer granular with an increasing value of $th_{sim}$.
For $th_{sim}=0$ (the least restrictive value), we observe that all units in the considered area are assigned to a single subgraph.
Note, that Algorithm \ref{alg:mergeSubrahps} with $th_{sim}=0$ will not automatically merge all subgraphs, but only those who share at least one common unit.
For $th_{sim}=0.1$ the junction is partitioned into three major subgraphs corresponding to the north (green), south (yellow) and west (purple) part of the network.  
Further increase of $th_{sim}$ leads to finer granular partitions of the junction.
For instance, for $th_{sim}=0.3$ individual subgraphs for roads in the north-west (pink) and the northeast (purple) are present. 
Finally, for $th_{sim} \in \{0.4,0.5\}$ we observe a fine granular partitioning of the junction in a large number of subgraphs, where individual subgraphs may contain only a few units. This corresponds to the rise of the number of subgraphs in Figure \ref{fig:subgraphSize} for high values of $th_{sim}$.

Overall, $th_{sim}$ can be used to adjust the granularity of \approach.
Whereas lower threshold values result in large subgraphs covering 
larger fractions of the road network ($th_{sim}=0$), subgraphs obtained using higher threshold values (e.g. $th_{sim}=0.4$) cover much smaller groups of affected units.

%% file: subgraphs.tex
\begin{figure}[!ht]
    \centering
    \begin{subfigure}[t]{0.32\textwidth}
        \includegraphics[width=\textwidth]{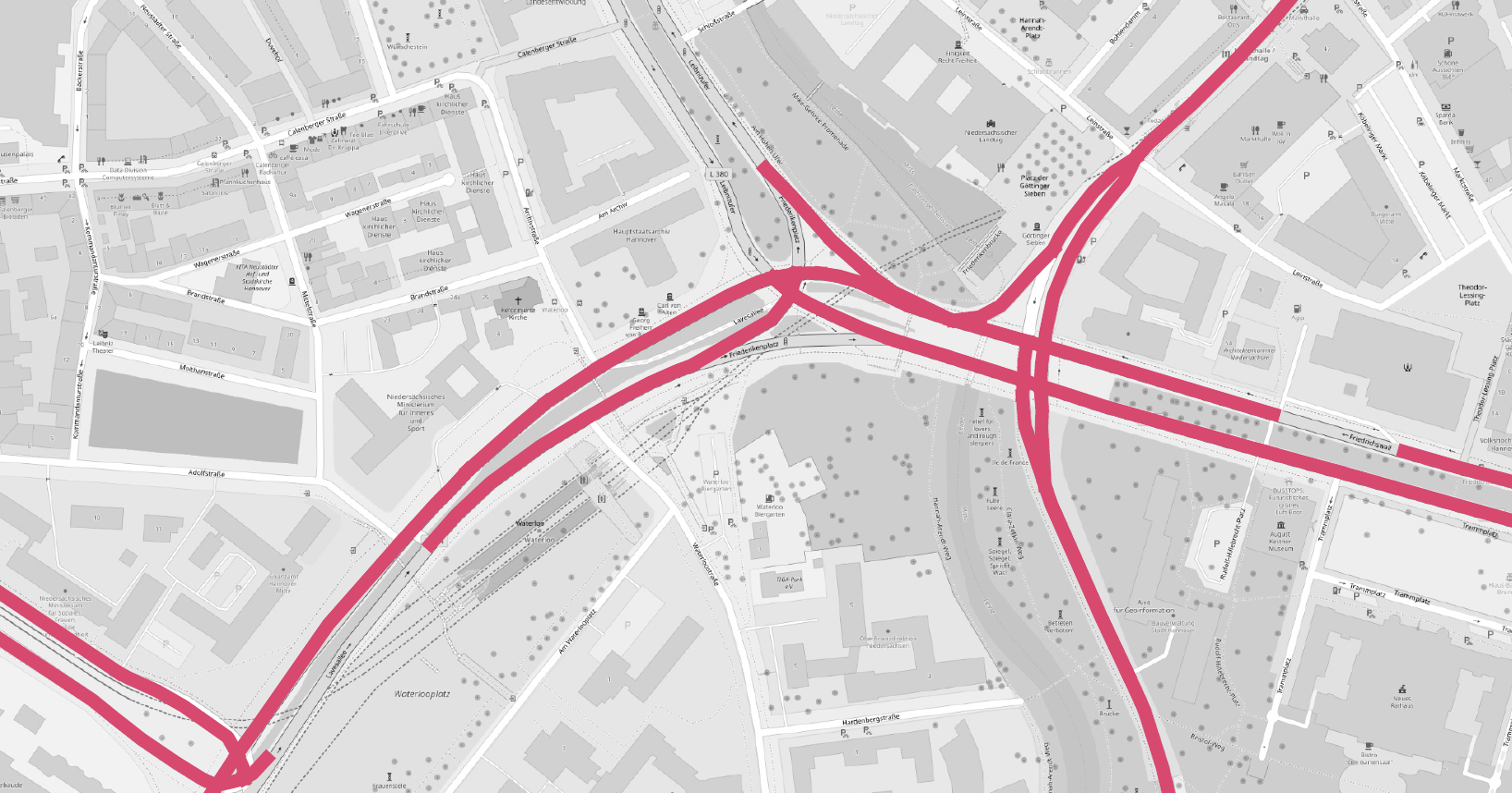}
        \caption{$th_{sim}=0.0$}
        \label{fig:subgraphs_00}
    \end{subfigure}
    \begin{subfigure}[t]{0.32\textwidth}
        \includegraphics[width=\textwidth]{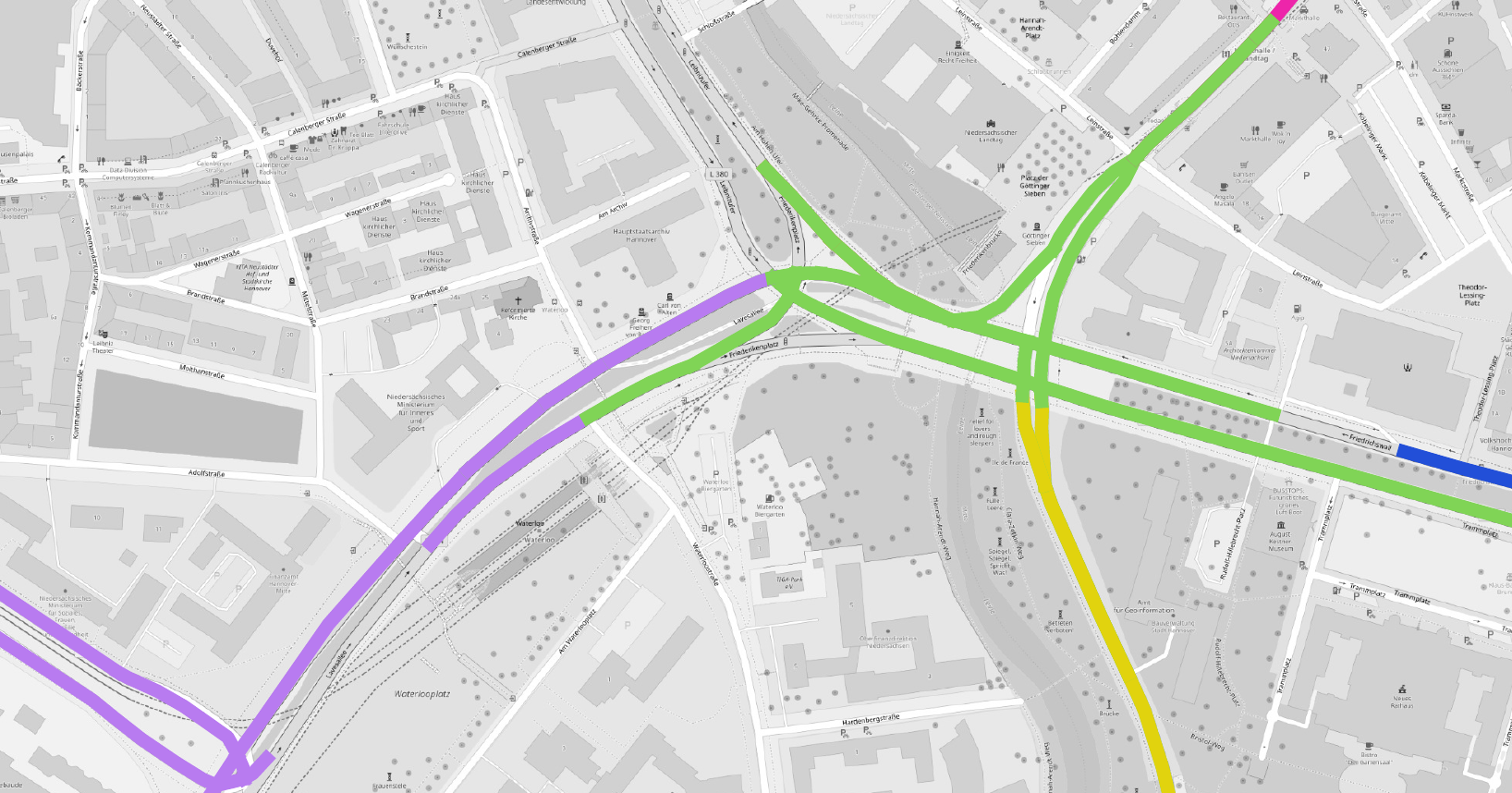}
        \caption{$th_{sim}=0.1$}
        \label{fig:subgraphs_01}
    \end{subfigure}
        \begin{subfigure}[t]{0.32\textwidth}
        \includegraphics[width=\textwidth]{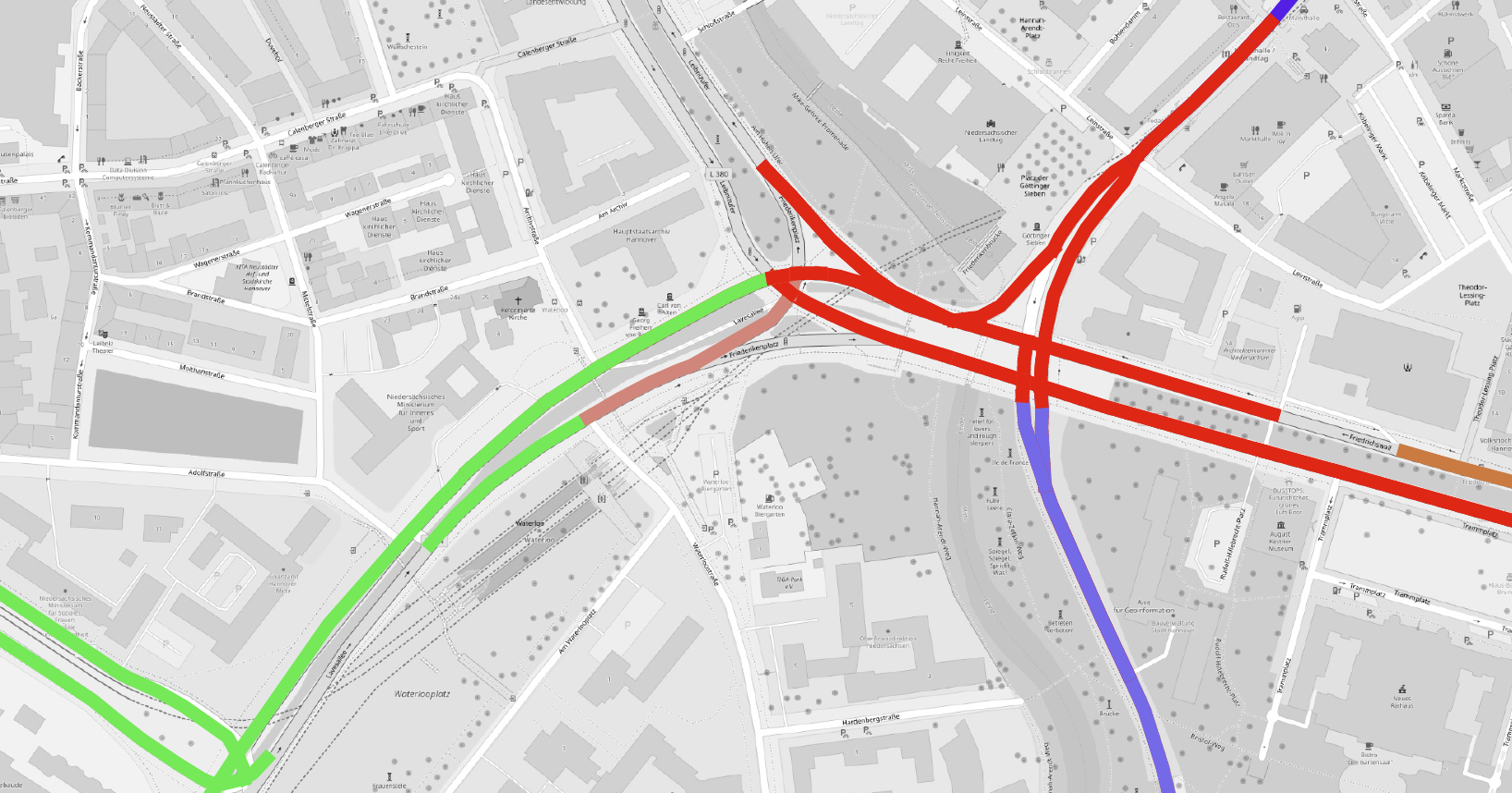}
        \caption{$th_{sim}=0.2$}
        \label{fig:subgraphs_02}
    \end{subfigure}\vspace{0.2cm}
    \begin{subfigure}[t]{0.32\textwidth}
        \includegraphics[width=\textwidth]{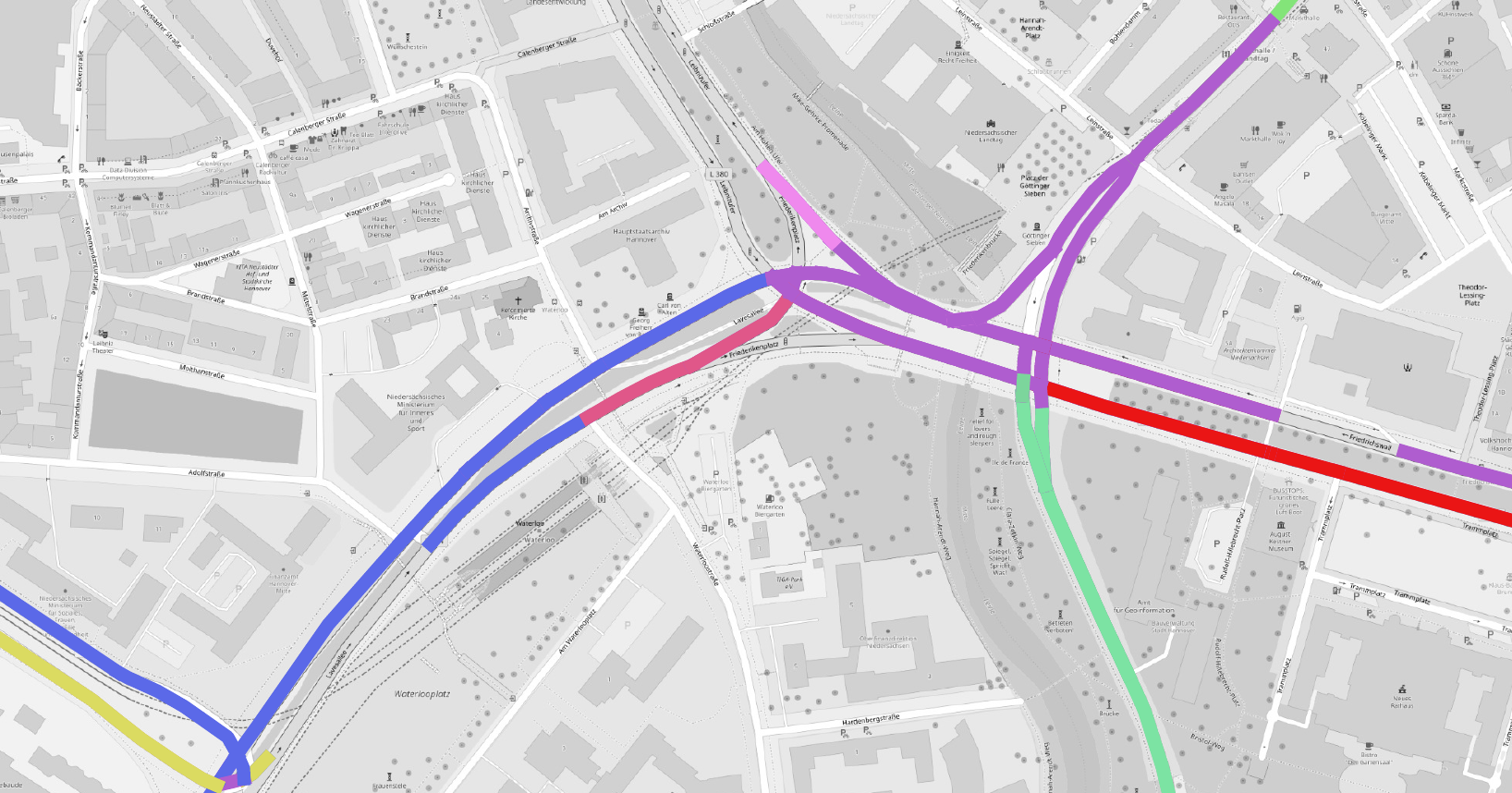}
        \caption{$th_{sim}=0.3$}
        \label{fig:subgraphs_03}
    \end{subfigure}
    \begin{subfigure}[t]{0.32\textwidth}
        \includegraphics[width=\textwidth]{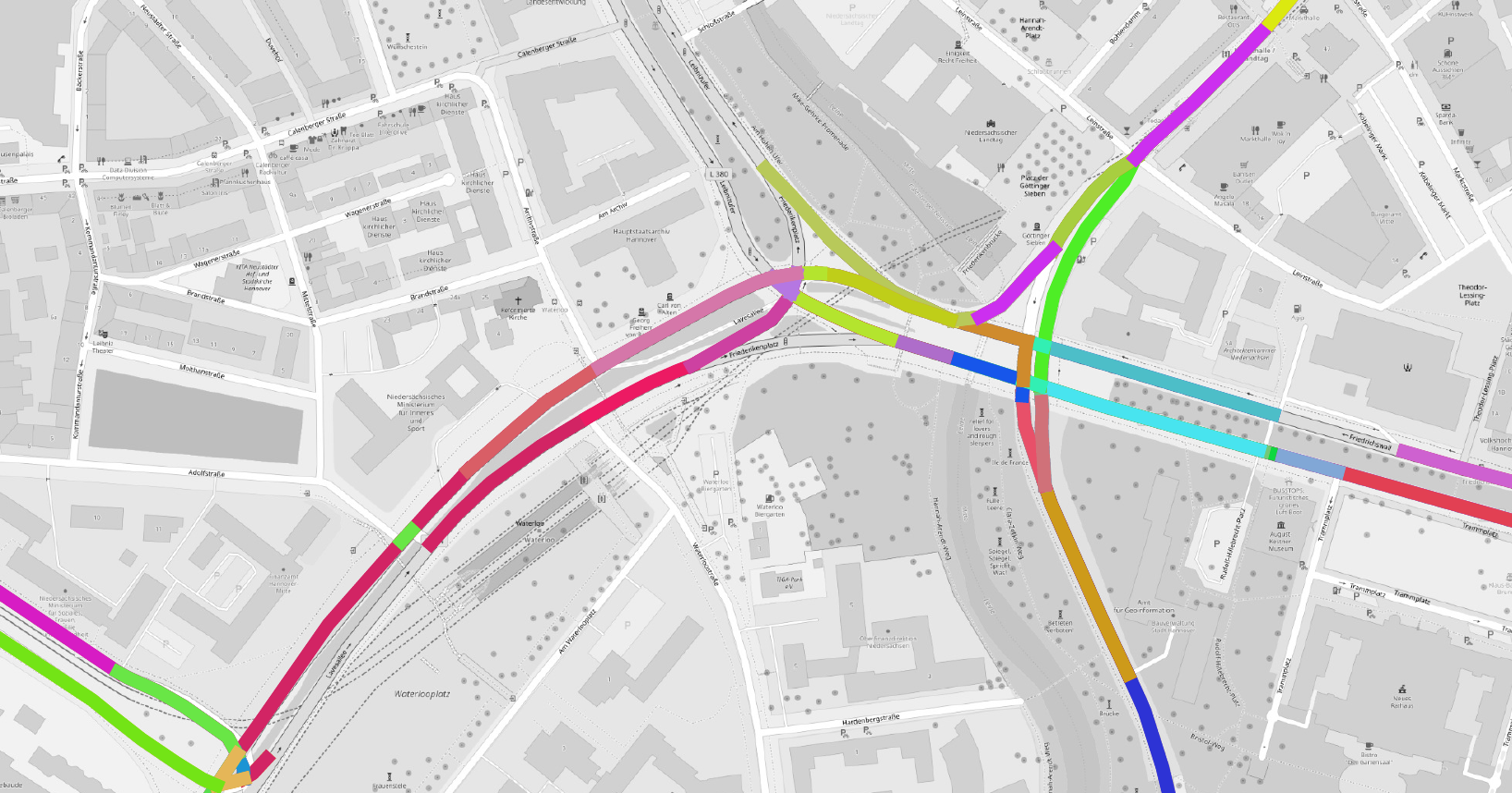}
        \caption{$th_{sim}=0.4$}
        \label{fig:subgraphs_04}
    \end{subfigure}
        \begin{subfigure}[t]{0.32\textwidth}
        \includegraphics[width=\textwidth]{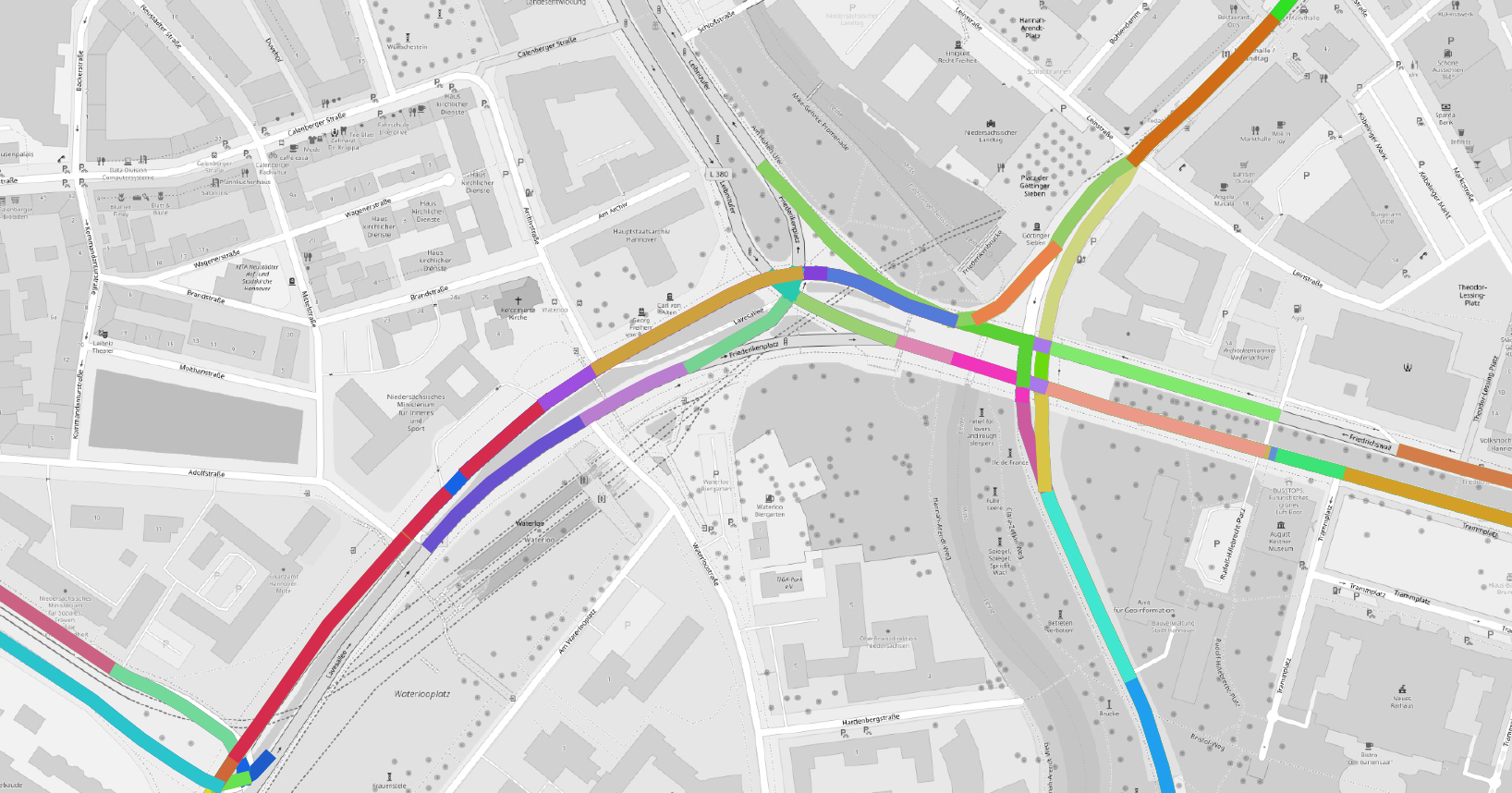}
        \caption{$th_{sim}=0.5$}
        \label{fig:subgraphs_05}
    \end{subfigure}
    \caption{Example of affected subgraphs calculated for a major junction in Hannover with varying values of $th_{sim}$. 
    Colours of the road segments indicate the subgraph assignment of the segments. Map images: \textcopyright OpenStreetMap contributors, ODbL.}
    \label{fig:subgraphs}
\end{figure}

%% file: 7_conclusion.tex
\section{Conclusion and Outlook}
\label{sec:conclusion}

In this article, we addressed the problem of data-driven discovery of topological dependencies of Recurrent Congestion within urban road networks.
We presented the \approach approach - a novel method to detect such dependencies based on traffic outlier analysis. 
\approach detects the units (i.e. road segments) within the road network that indicate an exceptionally high traffic load, proposes algorithms to identify affected subgraphs within the road network using these units and identifies spatio-temporal dependencies among these subgraphs. Furthermore, \approach provides parameters to adjust the granularity of the identified subgraphs to specific use cases.
Our evaluation results on the real-world datasets demonstrate that \approach can detect meaningful spatio-temporal dependencies among the subgraphs in urban road networks. The identified RC patterns include, for example, dependencies in the feeder roads of highways, alternative routes in case of traffic disruptions, or typical routes to POIs such as, e.g., event venues.
In future work, we intend to address the aspects of
explainability of \approach results for end-users, including, e.g., city planners and traffic managers.